%% file: main.tex
\documentclass[10pt,twocolumn,letterpaper]{article}

\usepackage[cvprpagenumbers]{cvpr}      
\usepackage{url}
\usepackage{multirow}
\usepackage{arydshln}
\usepackage{rotating}

\input{preamble}

\usepackage{ulem}

\definecolor{cvprblue}{rgb}{0.21,0.49,0.74}
\newcommand\blfootnote[1]{%
  \begingroup
  \renewcommand\thefootnote{}\footnote{#1}%
  \addtocounter{footnote}{-1}%
  \endgroup
}

\newcommand{\webpage}{{\url{https://rmurai.co.uk/projects/GaussianSplattingSLAM/}}}

\newcommand{\webvideo}{{\url{https://youtu.be/x604ghp9R_Q/}}}

\title{Gaussian Splatting SLAM\\
}

\author{
Hidenobu Matsuki$^{1}$$^{{*}}$ \qquad  Riku Murai$^{2}$$^{{*}}$ \qquad Paul H. J. Kelly$^{2}$ \qquad Andrew J. Davison$^{1}$ \vspace{0.5em}\\
$^{1}$Dyson Robotics Laboratory, Imperial College London \\  $^{2}$Software Performance Optimisation Group, Imperial College London \\
{\tt\small \{h.matsuki20, riku.murai15, p.kelly, a.davison\}@imperial.ac.uk}
\\ \\
\textbf{Website}: \mbox{{\webpage}} \\
\textbf{Video}: \mbox{{\webvideo}}
}

\begin{document}

\twocolumn[{
\renewcommand\twocolumn[1][]{#1}
\maketitle
\includegraphics[width=\linewidth]{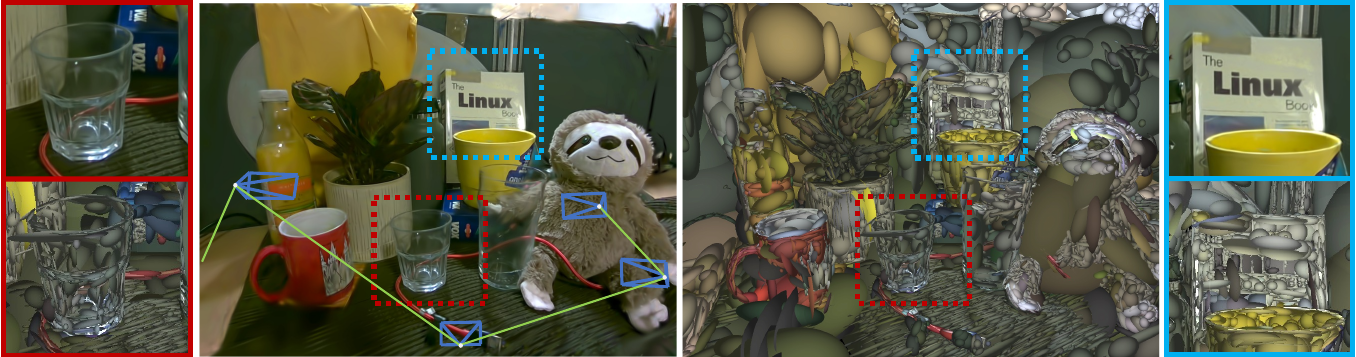}

\captionof{figure}{
From a single monocular camera, we reconstruct a high fidelity {\it3D scene} live at 3fps. For every incoming RGB frame, 3D Gaussians are incrementally formed and optimised together with the camera poses.
We show both the rasterised Gaussians (left) and Gaussians shaded to highlight the geometry (right). Notice the details and the complex material properties (e.g. transparency) captured. Thin structures such as wires are accurately represented by numerous small, elongated Gaussians, and transparent objects are effectively represented by placing the Gaussians along the rim. Our system significantly advances the fidelity a live monocular SLAM system can capture.
\vspace{2em}
}
\label{fig:teaser}
}]

\label{sec:intro}
\maketitle
{\blfootnote{*Authors contributed equally to this work.}}

\input{sec/0_abstract}    
\input{sec/1_intro}

\input{sec/2_relatedworks}
\input{sec/3_method}
\input{sec/4_evaluation}
\input{sec/5_conclusion}

\input{sec/6_acknowledgement}

\input{sec/X_suppl}

\normalem

\clearpage

{
    \small
    \bibliographystyle{ieeenat_fullname}
    \bibliography{main,robotvision}
}

\end{document}

%% file: preamble.tex
\usepackage[dvipsnames]{xcolor}

\DeclareMathOperator*{\argmin}{arg\,min}

%% file: sec/0_abstract.tex
\begin{abstract}
We present the first application of 3D Gaussian Splatting in monocular SLAM, the most fundamental but the hardest setup for Visual SLAM. Our method, which runs live at 3fps, utilises Gaussians as the only 3D representation, unifying the required representation for accurate, efficient tracking, mapping, and high-quality rendering. Designed for challenging monocular settings, our approach is seamlessly extendable to RGB-D SLAM when an external depth sensor is available. Several innovations are required to continuously reconstruct 3D scenes with high fidelity from a live camera. First, to move beyond the original 3DGS algorithm, which requires accurate poses from an offline Structure from Motion (SfM) system, we formulate camera tracking for 3DGS using direct optimisation against the 3D Gaussians, and show that this enables fast and robust tracking with a wide basin of convergence. Second, by utilising the explicit nature of the Gaussians, we introduce geometric verification and regularisation to handle the ambiguities occurring in incremental 3D dense reconstruction. Finally, we introduce a full SLAM system which not only achieves state-of-the-art results in novel view synthesis and trajectory estimation but also reconstruction of tiny and even transparent objects.
\end{abstract}

%% file: sec/1_intro.tex
\section{Introduction}
A long-term goal of online reconstruction with a single moving camera is
 near-photorealistic fidelity, which will surely allow new levels of performance in many areas of Spatial AI and robotics as well as opening up a whole range of new applications.
While we increasingly see the benefit of applying powerful pre-trained priors to 3D reconstruction, a key avenue for progress is still the invention and development of core 3D representations with advantageous properties. 
Many ``layered'' SLAM methods exist which tackle the SLAM problem by integrating multiple different 3D representations or existing SLAM components; however, the most interesting advances are when a new unified dense representation can be used for all aspects of a system's operation: local representation of detail, large-scale geometric mapping and also camera tracking by direct alignment. 

In this paper, we present the first online visual SLAM system based solely on the 3D Gaussian Splatting (3DGS) representation~\cite{kerbl3Dgaussians} recently making a big impact in offline scene reconstruction. In 3DGS a scene is represented by a large number of Gaussian blobs with orientation, elongation, colour and opacity. Other previous world/map-centric scene representations used for visual SLAM include occupancy or Signed Distance Function (SDF) voxel grids~\cite{Newcombe:etal:ISMAR2011}; meshes~\cite{surfelmeshing}; point or surfel clouds~\cite{Keller:etal:3DV2013, badslam}; and recently neural fields~\cite{Sucar:etal:ICCV2021}. Each of these has disadvantages: grids use significant memory and have bounded resolution, and even if octrees or hashing allow more efficiency they cannot be flexibly warped for large corrections~\cite{Vespa:etal:RAL2018, Niessner:etal:SIGGRAPH2013}; meshes require difficult, irregular topology to fuse new information; 
surfel clouds are discontinuous and difficult to fuse and optimise; 
and neural fields require expensive per-pixel raycasting to render. We show that 3DGS has none of these weaknesses. As a SLAM representation, it is most similar to point and surfel clouds, and inherits their efficiency, locality and ability to be easily warped or modified. However, it also represents geometry in a smooth, continuously differentiable way: a dense cloud of Gaussians merge together and jointly define a continuous volumetric function. And crucially, the design of modern graphics cards means that a large number of Gaussians can be efficiently rendered via ``splatting'' rasterisation, up to 200fps at 1080p. This rapid, differentiable rendering is integral to the tracking and map optimisation loops in our system.  

The 3DGS representation has up until now only been used in offline systems for 3D reconstruction with known camera poses, and we present several innovations to enable online SLAM.
We first derive the analytic Jacobian on Lie group of camera pose with respect to a 3D Gaussians map, and show that this can be seamlessly integrated into the existing differentiable rasterisation pipeline to enable camera poses to be optimised alongside scene geometry.
Second, we introduce a novel Gaussian isotropic shape regularisation to ensure geometric consistency, which we have found is important for incremental reconstruction.
Third, we propose a novel Gaussian resource allocation and pruning method to keep the geometry clean and enable accurate camera tracking. Our experimental results demonstrate photorealistic online local scene reconstruction, as well as state-of-the-art camera trajectory estimation and mapping for larger scenes compared to other rendering-based SLAM methods. We further show the uniqueness of the Gaussian-based SLAM method such as an extremely large camera pose convergence basin, which can also be useful for map-based camera localisation. Our method works with only monocular input, one of the most challenging scenarios in SLAM. To highlight the intrinsic capability of 3D Gaussian for camera localisation, our method does not use any pre-trained monocular depth predictor or other existing tracking modules, but relies solely on RGB image inputs in line with the original 3DGS. Since this is one of the most challenging SLAM scenario, we also show our method can easily be extended to RGB-D SLAM when depth measurements are available. 

In summary, our contributions are as follows:
\begin{itemize}
      \item The first near real-time SLAM system which works with a 3DGS as the only underlying scene representation, which can handle monocular only inputs.
      \item Novel techniques within the SLAM framework, including the analytic Jacobian on Lie group for direct camera pose estimation, isotropic regularisation of the Gaussian shape, and geometric verification.
      \item Extensive evaluations on a variety of datasets both for monocular and RGB-D settings, demonstrating competitive performance, particularly in real-world scenarios.
\end{itemize}

%% file: sec/2_relatedworks.tex
\section{Related Work}
\label{sec:formatting}

\textbf{Dense SLAM:}
Dense visual SLAM focuses on reconstructing detailed 3D maps, unlike sparse SLAM methods which excel in pose estimation ~\cite{Mur-Artal:etal:TRO2015, Engel:etal:PAMI2017, Forster:etal:ICRA2014} but typically yield maps useful mainly for localisation. In contrast, dense SLAM creates interactive maps beneficial for broader applications, including AR and robotics.
Dense SLAM methods are generally divided into two primary categories: Frame-centric and Map-centric. \textbf{Frame-centric SLAM} minimises photometric error across consecutive frames, jointly estimating per-frame depth and frame-to-frame camera motion.
Frame-centric approaches~\cite{Teed:Deng:NIPS2021, Czarnowski:etal:RAL2020} are efficient, as individual frames host local rather than global geometry (e.g. depth maps), 
and are attractive for long-session SLAM, but
if a dense global map is needed, it must be constructed on demand by assembling all of these parts which are not necessarily fully consistent.
In contrast, \textbf{Map-centric SLAM} uses a unified 3D representation across the SLAM pipeline, enabling a compact and streamlined system. Compared to purely local frame-to-frame tracking, a map-centric approach leverages global information by tracking against the reconstructed 3D consistent map. 
Classical map-centric approaches often use voxel grids  ~\cite{Newcombe:etal:ISMAR2011, Dai:etal:ACMTOG2017, Whelan:etal:IJRR2015, Prisacariu:etal:CORR2014} or points~\cite{Keller:etal:3DV2013,Whelan:etal:RSS2015,badslam} as the underlying 3D representation. 
While voxels enable a fast look-up of features in 3D, the representation is expensive, and the fixed voxel resolution and distribution are problematic when the spatial characteristics of the environment are not known in advance.
On the other hand, a  point-based map representation, such as surfel clouds, enables adaptive changes in resolution and spatial distribution by dynamic allocation of point primitives in the 3D space. 
Such flexibility benefits online applications such as SLAM with deformation-based loop closure~\cite{Whelan:etal:RSS2015, badslam}. However, optimising the representation to capture high fidelity is challenging due to the lack of correlation among the primitives.
Recently, in addition to classical graphic primitives, neural network-based map representations are a promising alternative. iMAP~\cite{Sucar:etal:ICCV2021} demonstrated the interesting properties of neural representation, such as sensible hole filling of unobserved geometry.
Many recent approaches combine the classical and neural representations to capture finer details~\cite{Zhu2022CVPR, Sandström2023ICCV, johari-et-al-2023, zhu2023nicer}; however, the large amount of computation required for neural rendering makes the live operation of such systems challenging.

\textbf{Differentiable Rendering:} 
The classical method for creating a 3D representation was to unproject 2D observations into 3D space and to fuse them via weighted averaging~\cite{Newcombe:etal:ISMAR2011, McCormac:etal:ICRA2017}. Such an averaging scheme suffers from over-smooth representation and lacks the expressiveness to capture high-quality details.
To capture a scene with photo-realistic quality, differentiable volumetric rendering~\cite{DVR} has recently been popularised with Neural Radiance Fields (NeRF)~\cite{Mildenhall:etal:ECCV2020}. Using a single Multi-Layer Perceptron (MLP) as a scene representation, NeRF performs volume rendering by marching along pixel rays, querying the MLP for opacity and colour. 
Since volume rendering is naturally differentiable, the MLP representation is optimised to minimise the rendering loss using multiview information to achieve high-quality novel view synthesis.
The main weakness of NeRF is its training speed. Recent developments have introduced explicit volume structures such as multi-resolution voxel grids~\cite{yu2022plenoxels,SunSC22,liu2020neural} or hash functions~\cite{mueller2022instant} to improve performance. 
Interestingly, these projects demonstrate that the main contributor to high-quality novel view synthesis is not the neural network but rather differentiable volumetric rendering, and that it is possible to avoid the use of an MLP and yet achieve comparable rendering quality to NeRF~\cite{yu2022plenoxels}.
However, even in these systems, per-pixel ray marching remains a significant bottleneck for rendering speed.
This issue is particularly critical in SLAM, where immediate interaction with the map is essential for tracking. In contrast to NeRF, 3DGS performs differentiable rasterisation. Similar to regular graphics rasterisations, by iterating over the primitives to be rasterised rather than marching along rays, 3DGS leverages the natural sparsity of a 3D scene and achieves a representation which is expressive to capture high-fidelity 3D scenes while offering significantly faster rendering. Several works have applied 3D Gaussians and differentiable rendering to static scene capture~\cite{keselman2022fuzzy, wang2022voge}, and in particular more recent works utilise 3DGS and demonstrate superior results in vision tasks such as dynamic scene capture~\cite{luiten2023dynamic, yang2023gs4d, wu20234d} and 3D generation~\cite{tang2023dreamgaussian, GaussianDreamer}.
Our method adopts a Map-centric approach, utilising 3D Gaussians as the only SLAM representation. Similar to surfel-based SLAM, we dynamically allocate the 3D Gaussians, enabling us to model an arbitrary spatial distribution in the scene. Unlike other methods such as ElasticFusion~\cite{Whelan:etal:RSS2015} and PointFusion~\cite{Keller:etal:3DV2013}, however, by using differentiable rasterisation, our SLAM system can capture high-fidelity scene details and represent challenging object properties by direct optimisation against information from every pixel.

%% file: sec/3_method.tex
\newcommand{\project}[0]{\pi}
\newcommand{\meanW}[0]{\boldsymbol{\mu}_{W}}
\newcommand{\meanC}[0]{\boldsymbol{\mu}_{C}}
\newcommand{\meanI}[0]{\boldsymbol{\mu}_{I}}
\newcommand{\covW}[0]{\boldsymbol{\Sigma}_{W}}
\newcommand{\covI}[0]{\boldsymbol{\Sigma}_{I}}
\newcommand{\cam}[0]{\boldsymbol{T}}
\newcommand{\camCW}[0]{\cam_{CW}}
\newcommand{\camWC}[0]{\cam_{WC}}
\newcommand{\rotCW}[0]{\boldsymbol{R}_{CW}}
\newcommand{\rotWC}[0]{\boldsymbol{R}_{WC}}
\newcommand{\tauC}[0]{\tau}
\newcommand{\thetaC}[0]{\theta}
\newcommand{\Exp}[0]{\text{Exp}}
\newcommand{\Log}[0]{\text{Log}}
\newcommand{\loss}[0]{\mathcal{L}}
\newcommand{\meddepth}[0]{\tilde{\mathcal{D}}}
\newcommand{\gaussians}[0]{\mathcal{G}}
\newcommand{\window}[0]{\mathcal{W}}
\newcommand{\kfwindow}[0]{\mathcal{W}_k}
\newcommand{\randwindow}[0]{\mathcal{W}_r}
\newcommand{\pd}[2]{\frac{\partial {#1} }{\partial {#2} }}
\newcommand{\mpd}[2]{\frac{\mathcal{D} {#1}}{\mathcal{D} {#2}}}
\newcommand{\se}[1]{\mathfrak{se}(#1)}
\newcommand{\SE}[1]{\boldsymbol{SE}(#1)}
\newcommand{\identity}[0]{\boldsymbol{I}}
\newcommand{\vectorize}[1]{vec(#1)}
\newcommand{\kron}[0]{\otimes}
\newcommand{\RR}{\mathbb{R}}
\newcommand{\gtimage}[0]{\bar{I}}
\newcommand{\gtdepth}[0]{\bar{D}}
\newcommand{\matJ}[0]{\mathbf{J}}
\newcommand{\matW}[0]{\mathbf{W}}
\newcommand{\vecs}[0]{\mathbf{s}}
\section{Method}

\begin{figure*}[!tbp]
  \center
  \includegraphics[width=\linewidth]{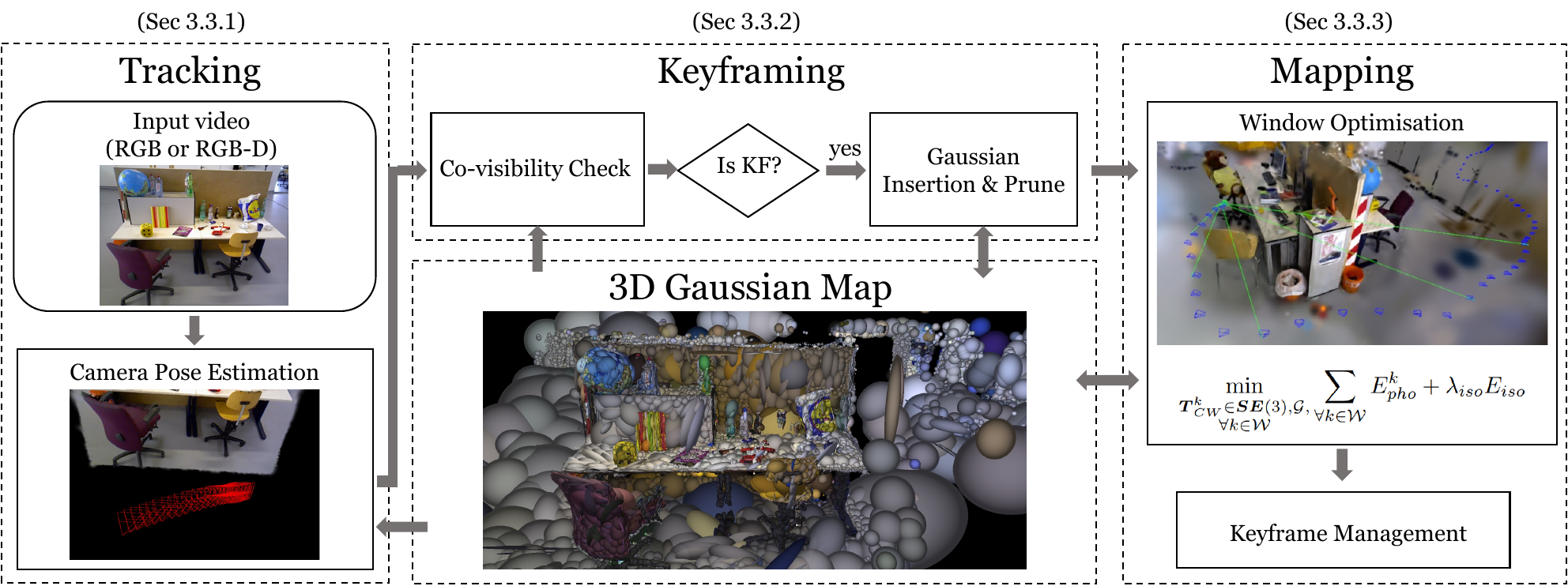}
  \caption{
  \textbf{SLAM System Overview:} 
Our SLAM system uses 3D Gaussians as the only representation, unifying all components of SLAM, including tracking, mapping, keyframe management, and novel view synthesis. 
  }\label{fig:system_diagram}
\end{figure*}

\subsection{Gaussian Splatting}
Our SLAM representation is 3DGS, mapping the scene with a set of anisotropic Gaussians $\gaussians$. Each Gaussian $\gaussians^i$ contains optical properties: colour $c^i$ and opacity $\alpha^i$.
For continuous 3D representation, the mean $\meanW^i$ and covariance  $\covW^i$, defined in the world coordinate, represent the Gaussian's position and its ellipsoidal shape. 
We omit the spherical harmonics (SHs) representing view-dependent radiance for simplicity but report the ablation with SHs in the supplementary.
Since 3DGS uses volume rendering, explicit extraction of the surface is not required. Instead, by splatting and blending $\mathcal{N}$ Gaussians, a pixel colour $\mathcal{C}_p$ is synthesised:
\begin{equation}
\mathcal{C}_p = \sum_{i\in\mathcal{N}} c_i \alpha_i \prod_{j=1}^{i-1} (1-\alpha_j)~.
\label{eqn:alpha_blending}
\end{equation}
3DGS performs rasterisation, iterating over the Gaussians rather than marching along the camera rays, and hence, free spaces are ignored during rendering.
During rasterisation, the contributions of $\alpha$ are decayed via a Gaussian function, based on the 2D Gaussian formed by splatting a 3D Gaussian. 
The 3D Gaussians $\mathcal{N}(\meanW, \covW)$ in world coordinates are related to the 2D Gaussians  $\mathcal{N}(\meanI, \covI)$ on the image plane through a projective transformation:
\begin{equation}
    \meanI = \project ( \camCW \cdot \meanW )~, \covI = \matJ\matW\covW \matW^T \matJ^T
    \label{eqn:mean_cov_w2i}
~,
\end{equation}
where $\project$ is the projection operation and $\camCW \in \SE{3}$ is the camera pose of the viewpoint. $\matJ$ is the Jacobian of the linear approximation of the projective transformation and $\matW$ is the rotational component of $\camCW$.
This formulation enables the 3D Gaussians to be differentiable and the blending operation provides gradient flow to the Gaussians. Using first-order gradient descent~\cite{Kingma:Ba:ICLR2015}, Gaussians gradually refines  both their optic and geometric parameters to represent the captured scene with high fidelity.
 
\subsection{Camera Pose Optimisation}
To achieve accurate tracking, we typically require at least 50 iterations of gradient descent per frame.
This requirement emphasises the necessity of a representation with computationally efficient view synthesis and gradient computation, making the choice of 3D representation a crucial part of designing a SLAM system.

In order to avoid the overhead of automatic differentiation, 3DGS implements rasterisation with CUDA with derivatives for all parameters calculated explicitly.
Since rasterisation is performance critical, we similarly derive the camera Jacobians explicitly.

To the best of our knowledge, we provide the first analytical Jacobian of $\SE{3}$ camera pose with respect to the 3D Gaussians used in EWA splatting~\cite{zwicker2002ewa} and 3DGS. This opens up new applications of 3DGS beyond SLAM.

We use Lie algebra to derive the minimal Jacobians, ensuring that the dimensionality of the Jacobians matches the degrees of freedom, eliminating any redundant computations.
The terms of Eq.~\eqref{eqn:mean_cov_w2i} are differentiable with respect to the camera pose $\camCW$; using the chain rule:
\begin{align}
    \pd{\meanI}{\camCW} &= \pd{\meanI}{\meanC}\mpd{\meanC}{\camCW}~, \label{eqn:grad_meani_cw}\\
    \pd{\covI}{\camCW} 
    &= \pd{\covI}{\matJ}\pd{\matJ}{\meanC} \mpd{\meanC}{\camCW}+ \pd{\covI}{\matW} \mpd{\matW}{\camCW} ~.  \label{eqn:grad_covi_cw}
\end{align}
where ${\camCW}$ represents the 3D position of Gaussian in the camera coordinate.
    We take the derivatives on the manifold to derive minimal parameterisation. Borrowing the notation from~\cite{Sola:etal:ARXIV2018}, let $\cam \in \SE{3}$ and $\tauC \in \se{3}$. We define the partial derivative on the manifold as:
\begin{equation}
    \mpd{f(\cam)}{\cam} \triangleq  \lim_{\tauC \to 0}\frac{\Log(f(\Exp(\tau) \circ \cam) \circ f(\cam)^{-1})}{\tauC}~,
\end{equation}
where $\circ$ is a group composition, and $\Exp, \Log$ are the exponential and logarithmic mappings between Lie algebra and Lie Group. With this, we derive the following:

\begin{align}
    \mpd{\meanC}{\camCW} =  \begin{bmatrix}\identity &-\meanC^\times\end{bmatrix},
    \mpd{\matW}{\camCW} = \begin{bmatrix}
    \mathbf{0} & -\matW_{:, 1}^\times \\
    \mathbf{0} & -\matW_{:, 2}^\times \\
    \mathbf{0} & -\matW_{:, 3}^\times\\
    \end{bmatrix}\label{eqn:grad_meanC_camCW_W_camCW}
    ~,
\end{align}
where ${}^{\times}$ denotes the skew symmetric matrix of a 3D vector, and $\matW_{:, i}$ refers to the $i$th column of the matrix.

\subsection{SLAM}\label{sec:SLAM}
In this section, we present details of full SLAM framework. The overview of the system is summarised in Fig.~\ref{fig:system_diagram}.
Please refer to the supplementary material for the further parameter details.\subsubsection{Tracking}\label{sec:tracking}

In tracking only the current camera pose is optimised, without updates to the map representation. 
In the monocular case, we minimise the following photometric residual:
\begin{equation}
    E_{pho} = \left\| I(\gaussians, \camCW) - \gtimage \right\|_1~,
    \label{eqn:photometric}
\end{equation}
where $I(\gaussians, \camCW)$ renders the Gaussians $\gaussians$ from $\camCW$, and $\gtimage$ is an observed image.

We further optimise affine brightness parameters for varying exposure and penalise non-edge or low-opacity pixels.
When depth observations are available, we define the geometric residual as:
\begin{equation}
    E_{geo} = \left\| D(\gaussians, \camCW) - \gtdepth \right\|_1~,
    \label{eqn:geometric_residual}
\end{equation}

where $D(\gaussians, \camCW)$ is depth rasterisation and $\gtdepth$ is the observed depth.
Rather than simply using the depth measurements to initialise the Gaussians, we minimise both photometric and geometric residuals: $\lambda_{pho} E_{pho} + (1 - \lambda_{pho}) E_{geo}$, where $\lambda_{pho}$ is a hyperparameter.

As in Eq.~\eqref{eqn:alpha_blending}, per-pixel depth is rasterised by alpha-blending:
\begin{equation}
\mathcal{D}_p = \sum_{i\in\mathcal{N}} z_i \alpha_i \prod_{j=1}^{i-1} (1-\alpha_j)~,
\label{eqn:alpha_blending_depth}
\end{equation}
where $z_i$ is the distance to the mean $\meanW$ of Gaussian $i$ along the camera ray. We derive analytical Jacobians for the camera pose optimisation in a similar manner to Eq.~\eqref{eqn:grad_meani_cw}, \eqref{eqn:grad_covi_cw}.

\subsubsection{Keyframing}~\label{sec:keyframing}
Since using all the images from a video stream to jointly optimise the Gaussians and camera poses online is infeasible, we maintain a small window $\kfwindow$ consisting of carefully selected keyframes based on inter-frame covisibility. Ideal keyframe management will select non-redundant keyframes observing the same area, spanning a wide baseline to provide better multiview constraints. The parameters are detailed in the supplementary.

\paragraph{Selection and Management}
Every tracked frame is checked for keyframe registration based on our simple yet effective criteria. We measure the covisibility by measuring the intersection over the union of the observed Gaussians between the current frame $i$ and the last keyframe $j$. If the covisibility drops below a threshold, or if the relative translation $t_{ij}$ is large with respect to the median depth, frame $i$ is registered as a keyframe.
For efficiency, we maintain only a small number of keyframes in the current window $\kfwindow$ following the keyframe management heuristics of DSO~\cite{Engel:etal:PAMI2017}. The main difference is that a keyframe is removed from the current window if the overlap coefficient with the latest keyframe drops below a threshold.

\paragraph{Gaussian Covisibility}\label{sec:gaussian_covisibility}
An accurate estimate of covisibility simplifies keyframe selection and management. 3DGS respects visibility ordering since the 3D Gaussians are sorted along the camera ray. This property is desirable for covisibility estimation as occlusions are handled by design.
A Gaussian is marked to be visible from a view if used in the rasterisation and if the ray's accumulated $\alpha$ has not yet reached 0.5. This enables our estimated covisibility to handle occlusions without requiring additional heuristics.

\paragraph{Gaussian Insertion and Pruning}\label{sec:GaussianInsertionPruning}
At every keyframe, new Gaussians are inserted into the scene to capture newly visible scene elements and to refine the fine details. When depth measurements are available, Gaussian means $\meanW$ are initialised by back-projecting the depth. In the monocular case, we render the depth at the current frame. For pixels with depth estimates, $\meanW$ are initialised around those depths with low variance; for pixels without the depth estimates, we initialise $\meanW$ around the median depth of the rendered image with high variance. 

In the monocular case, the positions of many newly inserted Gaussians are incorrect. While the majority will quickly vanish during optimisation as they violate multiview consistency, we further prune the excess Gaussians by checking the visibility amongst the current window $\kfwindow$. If the Gaussians inserted within the last 3 keyframes are unobserved by at least 3 other frames, we prune them out as they are geometrically unstable.

\subsubsection{Mapping}\label{sec:mapping}
\begin{figure}[!tbp]
  \center
  \includegraphics[width=\linewidth, trim={0 1.2cm 0 0},clip]{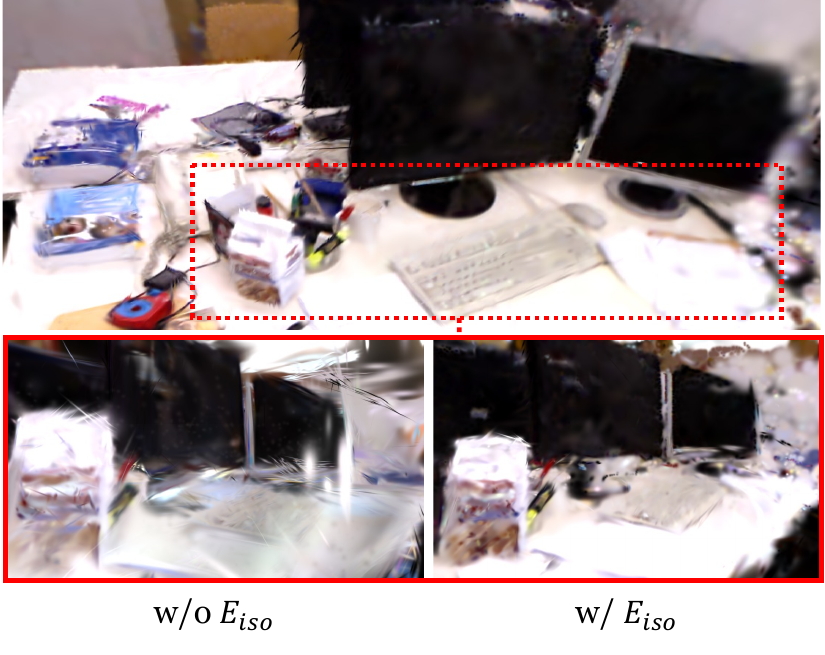}
  \caption{
   \textbf{Effect of isotropic regularisation}: \textbf{Top:} Rendering close to a training view (looking at the keyboard). \textbf{Bottom:} Rendering 3D Gaussians far from the training views (view from a side of the keyboard) without (left) and with (right) the isotropic loss.
When the photometric constraints are insufficient, the Gaussians tend to elongate along the viewing direction, creating artefacts in the novel views, and affecting the camera tracking.
  }\label{fig:isotropic_loss}
\end{figure}

The purpose of mapping is to maintain a coherent 3D structure and to optimise the newly inserted Gaussians.
During mapping, the keyframes in $\kfwindow$ are used to reconstruct currently visible regions. Additionally, two random past keyframes $\randwindow$ are selected per iteration to avoid forgetting the global map.
Rasterisation of 3DGS imposes no constraint on the Gaussians along the viewing ray direction, even with a depth observation. 
This is not a problem when sufficient carefully selected viewpoints are provided (e.g. in the novel view synthesis case); however, in continuous SLAM this causes many artefacts, making tracking challenging. We therefore introduce an isotropic regularisation:
\begin{equation}
    E_{iso} = \sum_{i =1}^{|\gaussians|} \left\| \vecs_i - \tilde{\vecs_i} \cdot \mathbf{1} \right\|_1
\end{equation}
to penalise the scaling parameters $\vecs_i$ (i.e. stretch of the ellipsoid) by its difference to the mean $\tilde{\vecs_i}$. As shown in Fig~\ref{fig:isotropic_loss}, this encourages sphericality, and avoids the problem of Gaussians which are highly elongated along the viewing direction creating artefacts.
Let the union of the keyframes in the current window and the randomly selected one be $\window = \kfwindow \cup \randwindow$. For mapping, we solve the following problem:
\begin{equation}
    \min_{\substack{\camCW^k \in \SE{3}, \gaussians, \\ \forall k \in \window}}
    \sum_{\forall k \in \window} E^{k}_{pho}+ \lambda_{iso} E_{iso}
    ~.\label{eqn:mapping}
\end{equation}
If depth observations are available, as in  tracking, geometric residuals Eq.~\eqref{eqn:geometric_residual} are added to the optimisation problem.

%% file: sec/4_evaluation.tex
\newcommand{\bhline}[1]{\noalign{\hrule height #1}}  

\section{Evaluation}
We conduct a comprehensive evaluation of our system across a range of both real and synthetic datasets. 
Additionally, we perform an ablation study to justify our design choices. Finally, we present qualitative results of our system operating live using a monocular camera,  illustrating its practicality and high fidelity reconstruction.

\subsection{Experimental Setup}

\paragraph{Datasets}
For our quantitative analysis, we evaluate our method on the TUM RGB-D dataset~\cite{Sturm:etal:IROS2012} (3 sequences) and the Replica dataset~\cite{replica19arxiv} (8 sequences), following the evaluation in \cite{Sucar:etal:ICCV2021}. For qualitative results, we use self-captured real-world sequences recorded by Intel Realsense d455. 
Since the Replica dataset is designed for RGB-D SLAM evaluation, it contains challenging purely rotational camera motions. We hence use the Replica dataset for RGB-D evaluation only. 
The TUM RGB-D dataset is used for both monocular and RGB-D evaluation.

\paragraph{Implementation Details}

We run our SLAM on a desktop with Intel Core i9 12900K 3.50GHz and a single NVIDIA GeForce RTX 4090. We present results from our multi-process implementation aimed at real-time applications. 
For a fair comparison with other methods on Replica, we additionally report result for single-process implementation which performs more mapping iterations.
As with 3DGS, time-critical rasterisation and gradient computation are implemented using CUDA. The rest of the SLAM pipeline is developed with PyTorch.
Details of hyperparameters are provided in the supplementary material.

\paragraph{Metrics}
For camera tracking accuracy, we report the Root Mean Square Error (RMSE) of the Absolute Trajectory Error (ATE) of the keyframes. To evaluate map quality, we report standard photometric rendering quality metrics (PSNR, SSIM and LPIPS) following the evaluation protocol used in~\cite{Sandström2023ICCV}. 
To evaluate the map quality, on every fifth frame, rendering metrics are computed. We exclude the keyframes (training views).
We report the average across three runs for all our evaluations. In the tables, the best result is in bold, and the second best is underlined.
 
\paragraph{Baseline Methods}
We primarily benchmark our SLAM method against other approaches that, like ours, do not have explicit loop closure. In monocular settings, we compare with state-of-the-art classical and learning-based direct visual odometry (VO) methods. Specifically, we compare DSO~\cite{Engel:etal:PAMI2017}, DepthCov~\cite{Dexheimer:etal:CVPR2023}, and DROID-SLAM~\cite{Teed:Deng:NIPS2021} in VO configurations. These methods are selected based on their public reporting of results on the benchmark (TUM dataset) or the availability of their source code for getting the benchmark result. Since one of our focuses is the online scale estimation under monocular scale ambiguity, the method which uses ground truth poses for the system initialisation such as ~\cite{li2023dense} is not considered for the comparison.
In the RGB-D case, we compare against neural-implicit SLAM methods~\cite{Sucar:etal:ICCV2021, Zhu2022CVPR, huang2021difusion, yang2022vox, johari-et-al-2023, wang2023coslam, Sandström2023ICCV} which are also map-centric, rendering-based and do not perform loop closure.

\subsection{Quantitative Evaluation}
\paragraph{Camera Tracking Accuracy}

Table~\ref{table:ate_tum_rgbd} shows the tracking results on the TUM RGB-D dataset. In the monocular setting, our method surpasses other baselines without requiring any deep priors. Furthermore, our performance is comparable to systems which perform explicit loop closure. This clearly highlights that there still remains potential for enhancing the tracking of monocular SLAM by exploring fundamental SLAM representations.

Our RGB-D method shows better performance than any other baseline method. Notably, our system surpasses ORB-SLAM in the fr1 sequences, narrowing the gap between Map-centric SLAM and the state-of-the-art sparse frame-centric methods.
Table~\ref{tab:ate_replica_rgbd} reports results on the synthetic Replica dataset. Our single-process implementation shows competitive performance and achieves the best result in 6 out of 8 sequences. Our multi-process implementation which performs fewer mapping iterations still performs comparably. 
In contrast to other methods, our system demonstrates higher performance on real-world data (TUM RGB-D), by optimising the Gaussian positions to compensate for the sensor noise.

\renewcommand{\arraystretch}{1.2}
\begin{table}[!t]
\centering
 \resizebox{\linewidth}{!}{
\begin{tabular}{ccccccc}
\hline
\textbf{Input} & {\begin{tabular}{c}\textbf{Loop-}\\\textbf{closure} \end{tabular}} & \textbf{Method} & \textbf{fr1/desk} & \textbf{fr2/xyz} & \textbf{fr3/office} & \textbf{Avg.} \\ \hline

 \multirow{6}{*}{\begin{turn}{90}Monocular\end{turn}}
 & \multirow{4}{*}{w/o} & DSO~\cite{Engel:etal:PAMI2017} & 22.4 & \bf1.10 & 9.50 & 11.0 \\ 
 && DROID-VO~\cite{Teed:Deng:NIPS2021} & \underline{5.20} & 10.7 & \underline{7.30} & \underline{7.73} \\
 && DepthCov-VO~\cite{Dexheimer:etal:CVPR2023} & 5.60 & \underline{1.20} & 68.8 & 25.2\\
 && \bf{Ours}   & \bf3.78 &  4.60 &  \bf3.50 &  \bf3.96 \\ \cline{2-7}\noalign{\vskip\doublerulesep
         \vskip-\arrayrulewidth} \cline{2-7}
    
 &\multirow{2}{*}{w/} & DROID-SLAM~\cite{Teed:Deng:NIPS2021} & \bf{1.80} & \bf{0.50} & 2.80 & 1.70 \\
 && ORB-SLAM2~\cite{Mur-Artal:etal:TRO2017} & 1.90 & 0.60 & \bf{2.40} & \bf{1.60} \\
 \hline \hline

\multirow{11}{*}{\begin{turn}{90}RGB-D\end{turn}} 
& \multirow{8}{*}{w/o} & iMAP~\cite{Sucar:etal:ICCV2021} & 4.90 & 2.00 & 5.80 & 4.23 \\ 
 && NICE-SLAM~\cite{Zhu2022CVPR} & 4.26 & 6.19 & 3.87 & 4.77 \\ 
 && DI-Fusion~\cite{huang2021difusion} & 4.40 & 2.00 & 5.80 & 4.07 \\ 
 && Vox-Fusion~\cite{yang2022vox} & 3.52 & 1.49 & 26.01 & 10.34 \\
 && ESLAM~\cite{johari-et-al-2023} & 2.47 & \bf{1.11} & 2.42 & \underline{2.00} \\ 
 && Co-SLAM~\cite{wang2023coslam} & \underline{2.40} & 1.70 & \underline{2.40} & 2.17 \\ 
 && Point-SLAM~\cite{Sandström2023ICCV} & 4.34 & \underline{1.31} & 3.48 & 3.04 \\
 && \textbf{Ours} & \textbf{1.50} & 1.44 & \textbf{1.49} & \textbf{1.47} \\ \cline{2-7}\noalign{\vskip\doublerulesep
         \vskip-\arrayrulewidth} \cline{2-7}

 & \multirow{3}{*}{w/}& BAD-SLAM~\cite{badslam} & 1.70 & 1.10 & 1.70 & 1.50 \\ 
 && Kintinous~\cite{Whelan:etal:IJRR2015} & 3.70 & 2.90 & 3.00 & 3.20 \\ 
 && ORB-SLAM2~\cite{Mur-Artal:etal:TRO2017} & \textbf{1.60} & \textbf{0.40} & \textbf{1.00} & \textbf{1.00} \\ 
\hline
\end{tabular}
}

\caption{\textbf{Camera tracking result on TUM for monocular and RGB-D.} ATE RMSE in cm is reported.
In both monocular and RGB-D cases, we achieve state-of-the-art performance. In particular, in the monocular case, not only do we outperform systems which use deep prior, but we achieve comparable performance with many of the RGB-D systems.
}
\label{table:ate_tum_rgbd}
\end{table}

\renewcommand{\arraystretch}{1.0}

\begin{table}[!t]
\centering
 \resizebox{\linewidth}{!}{
\begin{tabular}{ccccccccccc}
\hline
\textbf{Method} & \textbf{r0} & \textbf{r1} & \textbf{r2} & \textbf{o0} & \textbf{o1} & \textbf{o2} & \textbf{o3} & \textbf{o4} & \textbf{Avg.} \\ \hline
iMAP~\cite{Sucar:etal:ICCV2021}     & 3.12         & 2.54         & 2.31         & 1.69             & 1.03             & 3.99             & 4.05             & 1.93             & 2.58          \\
 \begin{tabular}{c} NICE-SLAM \\   \end{tabular}       & 0.97         & 1.31         & 1.07         & 0.88             & 1.00             & 1.06             & 1.10             & 1.13             & 1.07          \\
Vox-Fusion~\cite{yang2022vox}      & 1.37         & 4.70         & 1.47         & 8.48             & 2.04             & 2.58             & 1.11             & 2.94             & 3.09          \\ 
ESLAM~\cite{johari-et-al-2023}           & {0.71}         & 0.70         & 0.52         & {0.57}             & {0.55}             & 0.58             & 0.72             & \bf{0.63}             & 0.63          \\
Point-SLAM~\cite{Sandström2023ICCV}      & {0.61}         & {0.41}         & {0.37}         & \underline{0.38}             & \underline{0.48}             & {0.54}          & {0.69}             & \underline{0.72}             & \underline{0.53}          \\
\bf{Ours} & \underline{0.44}         &\underline{0.32}         & \underline{0.31}         &  {0.44}             & 0.52             & \bf{0.23}             & \underline{0.17}             & 2.25             & {0.58}          \\ 
{\begin{tabular}{c}\bf{Ours (sp)} \end{tabular}} & \bf{0.33}         & \bf{0.22}         & \bf{0.29}         & \bf{0.36}             & \bf{0.19}             & \underline{0.25}             & \bf{0.12}             & 0.81             & \bf{0.32}          \\ 

\hline
\end{tabular}
}
\caption{\textbf{Camera tracking result on Replica for RGB-D SLAM.} ATE RMSE in cm is reported. We achieve best performance across most sequences. Here, Ours is our multi-process implementation and Ours (sp) is the single-process implementation which ensures a certain amount of mapping iteration similar to other works. 
}
\label{tab:ate_replica_rgbd}
\end{table}

\begin{table}[!t]
\centering
\scalebox{0.7}{
\begin{tabular}{cccccc}
\hline
\textbf{Input} & \textbf{Method} & \textbf{fr1/desk} & \textbf{fr2/xyz} & \textbf{fr3/office} & \textbf{Avg.} \\ \hline

 \multirow{3}{*}{\begin{turn}{90}Mono\end{turn}}
 & w/o $E_{iso}$ & 4.16 &  4.66 & 5.73 & 4.83 \\ 
 & w/o kf selection & 13.2 & \bf{4.36} & 8.65 & 8.73  \\
 & \bf{Ours}   & \bf3.78 &  {4.60} &  \bf3.50 &  \bf3.96 \\ 
 \hline 
  \multirow{3}{*}{\begin{turn}{90}RGB-D\end{turn}}
 & w/o $E_{geo}$ & 2.39 &  \textbf{0.62} & 4.98 &   2.66 \\ 
 & w/o kf selection & 1.64 & 1.49  & 2.60  & 1.90   \\
 & \bf{Ours}  &  \textbf{1.50} & {1.44} & \textbf{1.49}  & \textbf{1.47}   \\ 
 \hline 
\end{tabular}
}

\caption{\textbf{Ablation Study on TUM RGB-D dataset.} We analyse the usefulness of isotropic regularisation, geometric residual, and keyframe selection to our SLAM system. Further isotropic regularisation ablation is available in supplementary.}
\label{table:ablation}
\end{table}

\begin{table}[!t]
\centering
 \resizebox{\linewidth}{!}{
\begin{tabular}{ccccc}
\hline
\multicolumn{5}{c}{\bf{Memory Usage [MB]}} \\
\hline
iMAP~\cite{Sucar:etal:ICCV2021} & NICE-SLAM~\cite{Zhu2022CVPR} 
& Co-SLAM~\cite{wang2023coslam}  & Ours (Mono) & Ours  (RGB-D) \\

\bf{0.8MB} & 40.3.4MB &  6.4MB & \underline{2.6MB}  & 3.97MB \\

 \hline 

\end{tabular}
}

\caption{\textbf{Memory Analysis on TUM RGB-D dataset.} The baseline numbers are computed from the parameter numbers in~\cite{wang2023coslam}
}
\label{table:time_memory}
\end{table}

\begin{figure}[!tbp]
  \center
  \includegraphics[width=\linewidth]{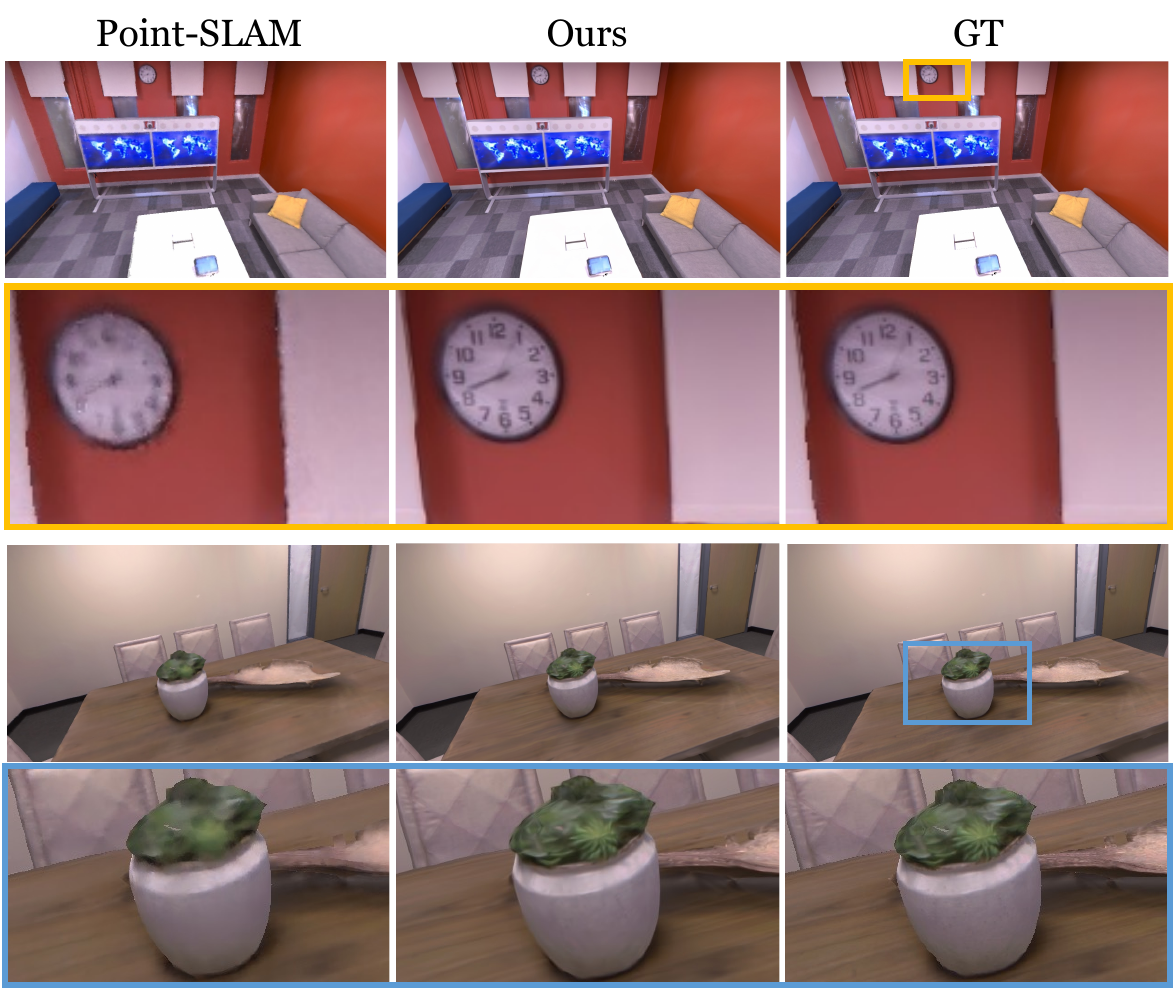}
  \caption{
   \textbf{Rendering examples on Replica.} Point-SLAM struggle with rendering fine details due to the stochastic ray sampling.
   }\label{fig:nvs}
\end{figure}

\paragraph{Novel View Rendering}
Table~\ref{table:rendering} summarises the novel view rendering performance of our method with RGB-D input. We consistently show the best performance across most sequences and is least second best. Our rendering FPS is hundreds of times faster than other methods, offering a significant advantage for applications which require real-time map interaction.
While Point-SLAM is competitive, that method focuses on view synthesis rather than novel-view synthesis. Their view synthesis is conditional on the availability of depth due to the depth-guided ray-sampling, making novel-view synthesis challenging. On the other hand, our rasterisation-based approach does not require depth guidance and achieves efficient, high-quality, novel view synthesis. Fig.~\ref{fig:nvs} provides a qualitative comparison of the rendering of ours and Point-SLAM (with depth guidance).

\renewcommand{\arraystretch}{1.0}
{
\begin{table}
\centering
 \resizebox{\linewidth}{!}{
\begin{tabular}{ccccc}
\hline
\textbf{Method}& \textbf{PSNR[db]↑}& \textbf{SSIM↑}& \textbf{LPIPS↓} & \textbf{Rendering FPS} \\
\hline
{\begin{tabular}{c}NICE-SLAM\cite{Zhu2022CVPR}\end{tabular}} &24.42 &0.809 &0.233 &0.54\\

{\begin{tabular}{c}Vox-Fusion\cite{yang2022vox}\end{tabular}}&24.41 &0.801 &0.236 &\underline{2.17} \\
{\begin{tabular}{c} Point-SLAM~\cite{Sandström2023ICCV} \end{tabular}} &\underline{35.17} &\bf{0.975} &0.124 &1.33  \\
\bf{ours} &\bf{38.94} &\underline{0.968} &\bf{0.070} &\bf{769} \\
\hline

\end{tabular}
}
\caption{\textbf{Average rendering performance on Replica (RGB-D).} Our method outperforms most of the rendering metrics compared to existing methods. Note that Point-SLAM uses ground-truth depth to guide sampling along rays. The full detail is available in supplementary.}
\label{table:rendering}
\end{table}
}

\renewcommand{\arraystretch}{1}

\paragraph{Ablative Analysis}
In Table~\ref{table:ablation}, we perform ablation to confirm our design choices.  Isotropic regularisation and geometric residual improve the tracking of monocular and RGB-D SLAM respectively, as they aid in constraining the geometry when photometric signals are weak.
For both cases, keyframe selection significantly improves systems performance, as it automatically chooses suitable keyframes based on our occlusion-aware keyframe selection and management.
We further compare the memory usage of different 3D representations in Table~\ref{table:time_memory}. MLP-based iMAP is clearly more memory efficient, but it struggles to express high-fidelity 3D scenes due to the limited capacity of small MLP. Compared with a voxel grid of features used in NICE-SLAM, our method uses significantly less memory. 

\paragraph{Convergence Basin Analysis}

\begin{figure}[!tbp]
  \center
  \includegraphics[width=\linewidth]{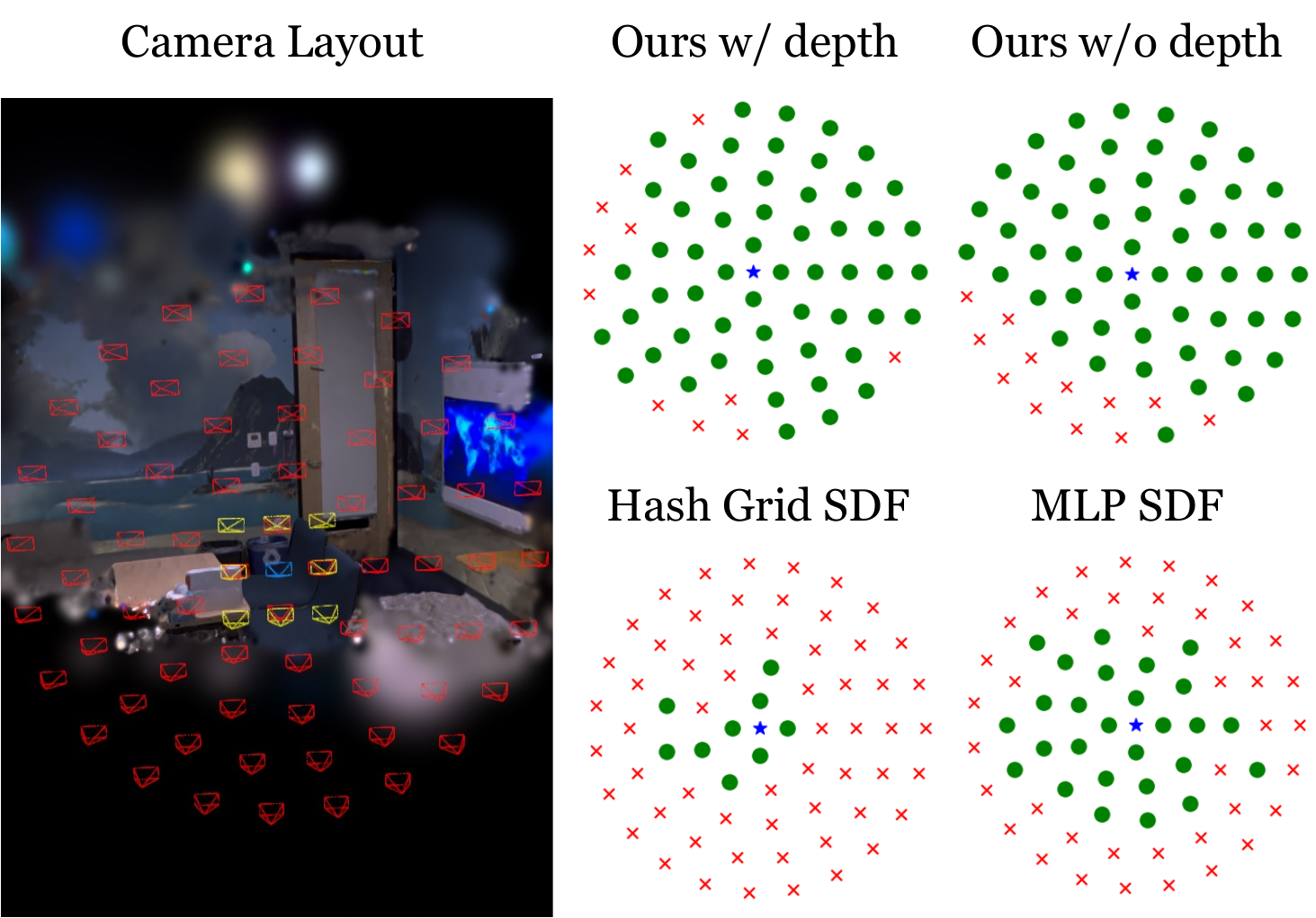}
  \caption{
   \textbf{\bf{Convergence basin analysis}}: \textbf{Left:} 3D Gaussian map from training views (Yellow) and visualisation of the test poses (Red) and target pose (Blue).
\textbf{Right:}  Convergence basin of our method. The green marks success, and the red marks failure.
   }\label{fig:conv_basin}
\end{figure}

\begin{figure}[!tbp]
  \includegraphics[width=\linewidth]{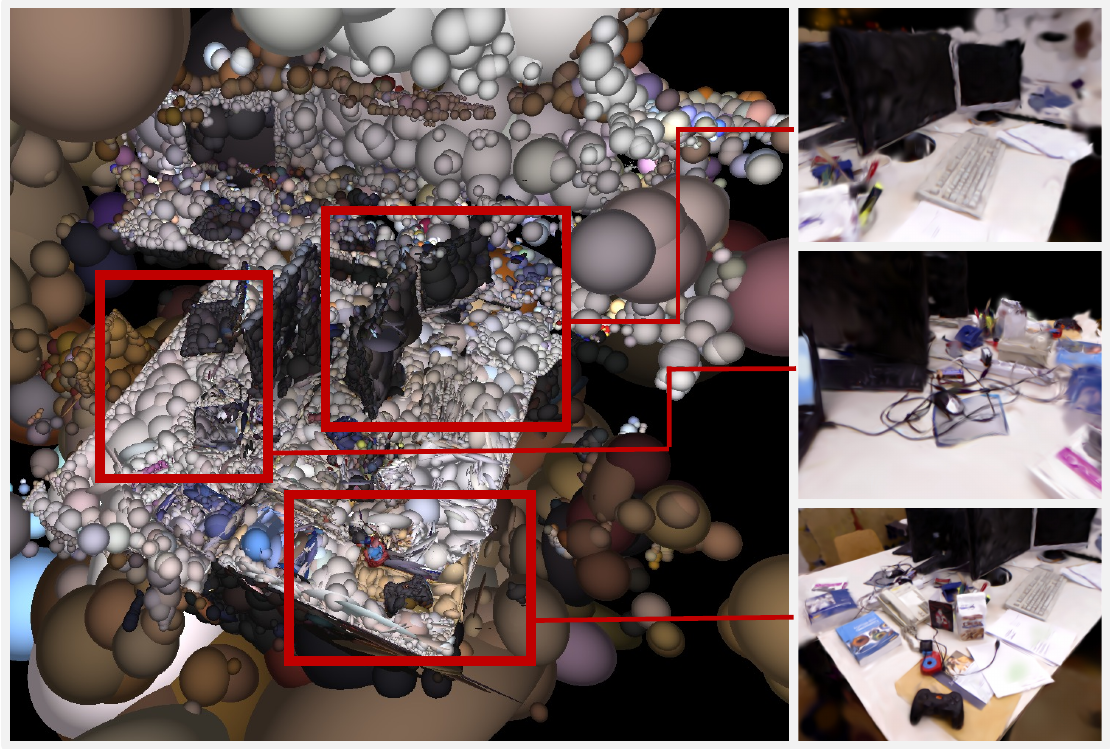}
  \caption{
  \textbf{Monocular SLAM result on fr1/desk sequence:} We show the reconstructed 3D Gaussian maps (Left) and novel view synthesis result (Right).}\label{fig:qualitative_tum}
\end{figure}

\begin{table}[h]
\centering
\scalebox{0.9}
 {
\begin{tabular}{ccccc}
\hline
\ \textbf{Method} & \textbf{seq1} & \textbf{seq2} & \textbf{seq3} & \textbf{Avg.} \\ \hline

   Neural SDF (Hash Grid)   & 0.13  & 0.15   & 0.16   & 0.14   \\ 
 Neural SDF (MLP)   &  0.40 & 0.38  & 0.22 & 0.33  \\ 

  Ours w/o depth & \underline{0.82}  & \underline{0.91} & \textbf{0.65} & \underline{0.79}  \\ 
  Ours w/ depth & \textbf{0.83} & \textbf{1.0} & \textbf{0.65} & \textbf{0.82}\\
 \hline 
\end{tabular}
}

\caption{\textbf{Camera convergence analysis.} We report the ratio of successful camera convergence for the different sequences, across different differentiable 3D representations.}
\label{table:conv_basin}
\end{table}

In our SLAM experiments, we discovered that 3D Gaussian maps have a notably large convergence basin for camera localisation. 
To investigate further, we conducted a convergence funnel analysis, an evaluation methodology proposed in ~\cite{Mitra:etal:GPG2004} and used in ~\cite{Newcombe:PHD2012}. Here, we train a 3D representation (e.g. 3DGS) using 9 fixed views arranged in a square. We set the viewpoint in the middle of the square to be the target view. As shown in Fig~\ref{fig:conv_basin}, we uniformly sample a position, creating a funnel. 
From the sampled position, given the RGB image of the target view, we perform camera pose optimisation for 1000 iterations. 
The optimisation is successful if it converges to within 1cm of the target view within the fixed iterations.
We compare our Gaussian approach with Co-SLAM~\cite{wang2023coslam}'s network (Hash Grid SDF) and iMAP's~\cite{Sucar:etal:ICCV2021} network with Co-SLAM's SDF loss for further geometric accuracy (MLP Neural SDF). We render the training views using a synthetic Replica dataset and create three sequences for testing (seq1, seq2 and seq3). The width of the square formed by the training view is 0.5m, and the test cameras are distributed with radii ranging from 0.2m to 1.2m, covering a larger area than the training view. When training the map, the three methods— Ours w/depth, Hash Grid SDF, and MLP SDF—use RGB-D images, whereas Ours w/o depth utilises only colour images. Fig.~\ref{fig:conv_basin} shows the qualitative results and Table~\ref{table:conv_basin} reports the success rate. For both with and without depth for training, our method shows better convergence. Unlike hashing and positional encoding which can lead to signal conflict, anisotropic Gaussians form a smooth gradient in 3D space, increasing the convergence basin. Further experimental details are available in the supplementary.

\subsection{Qualitative Results}
\begin{figure}[!tbp]
  \center
  \includegraphics[width=0.49\linewidth]{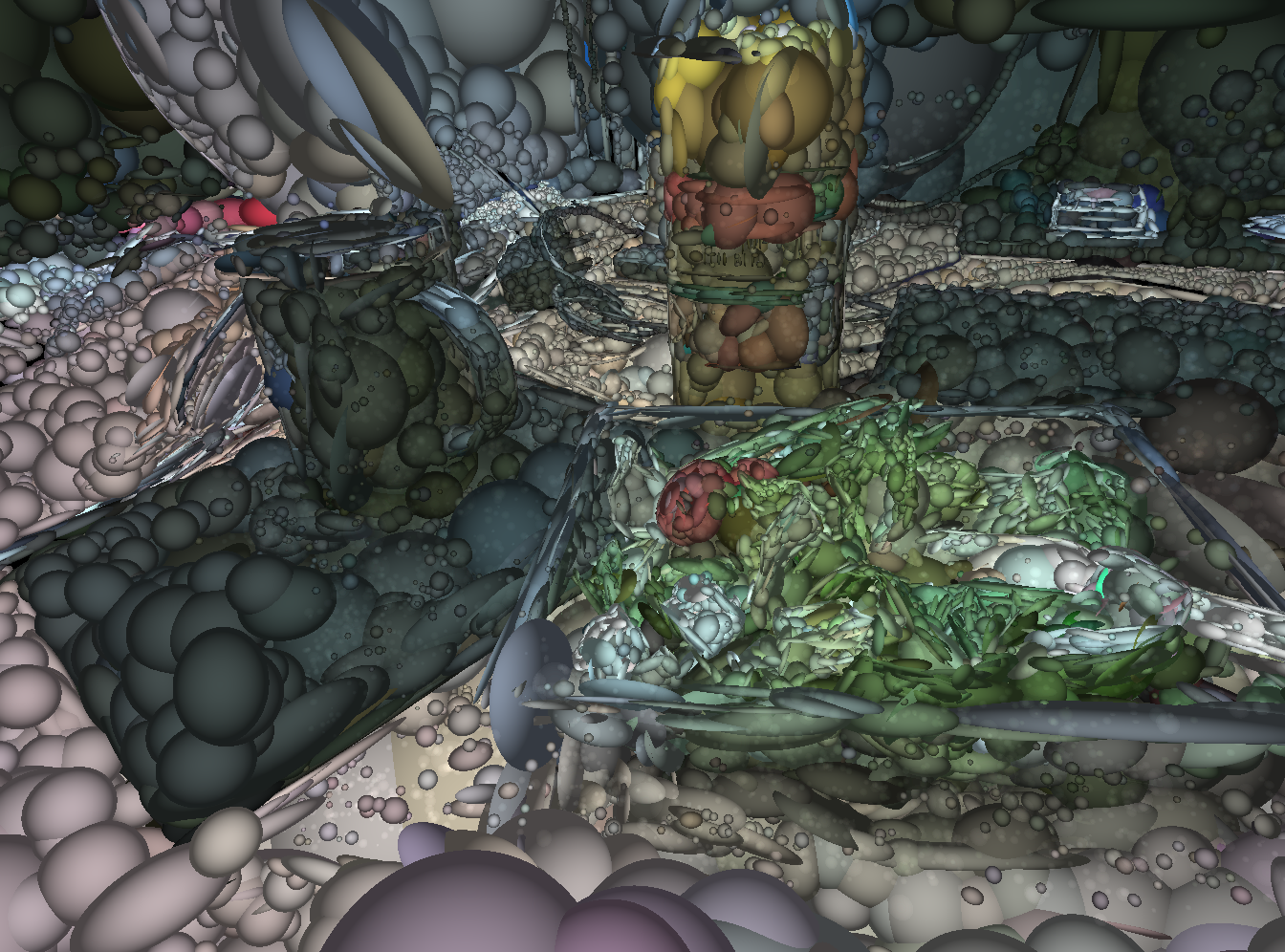}
  \includegraphics[width=0.49\linewidth]{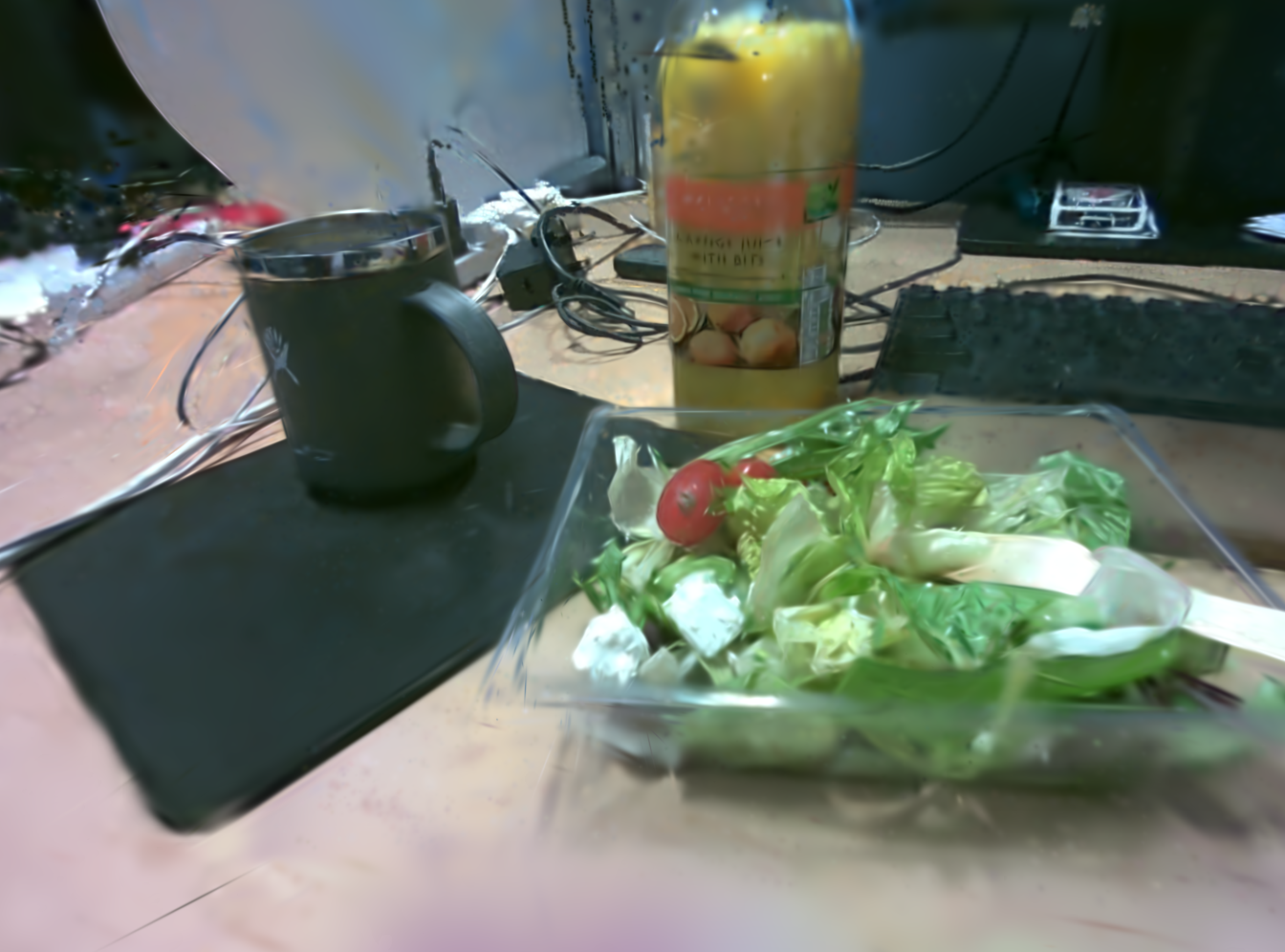}\\
    \includegraphics[width=0.49\linewidth]{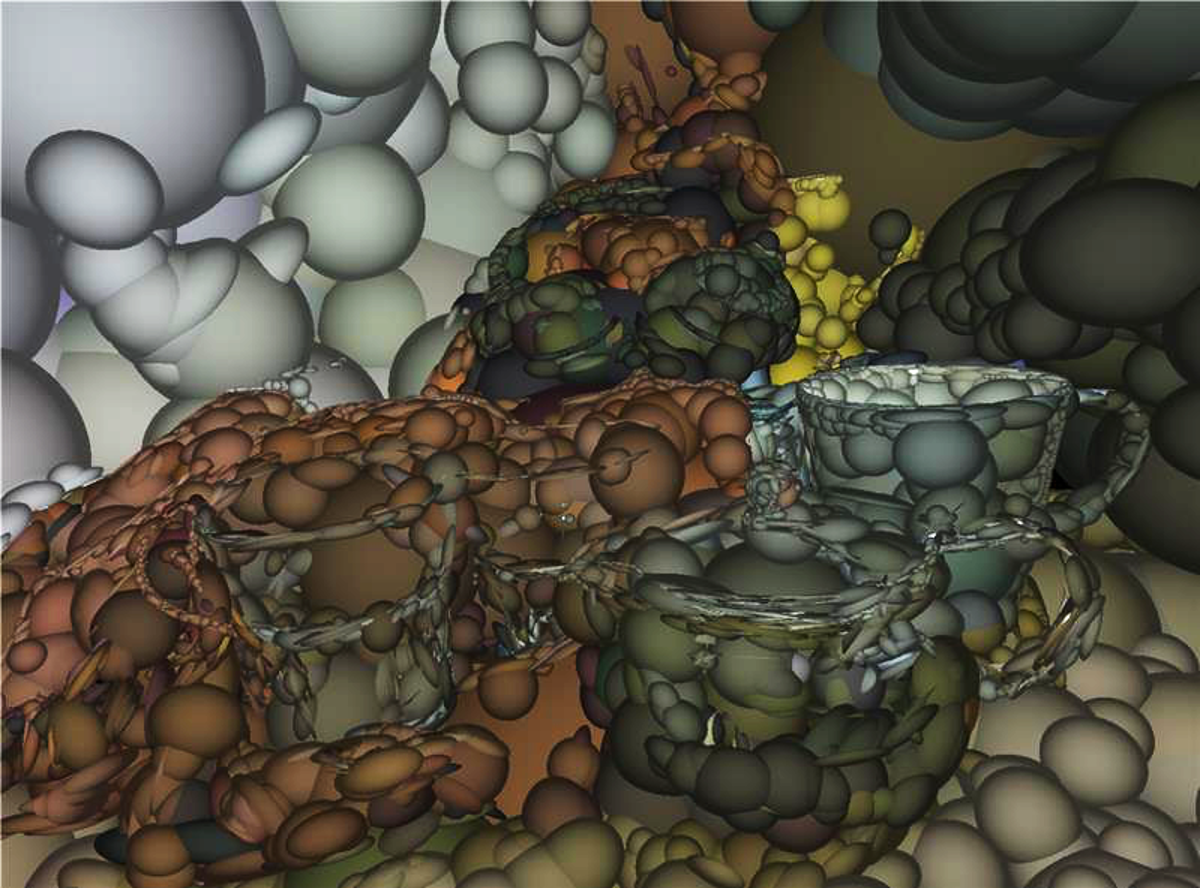}
  \includegraphics[width=0.49\linewidth]{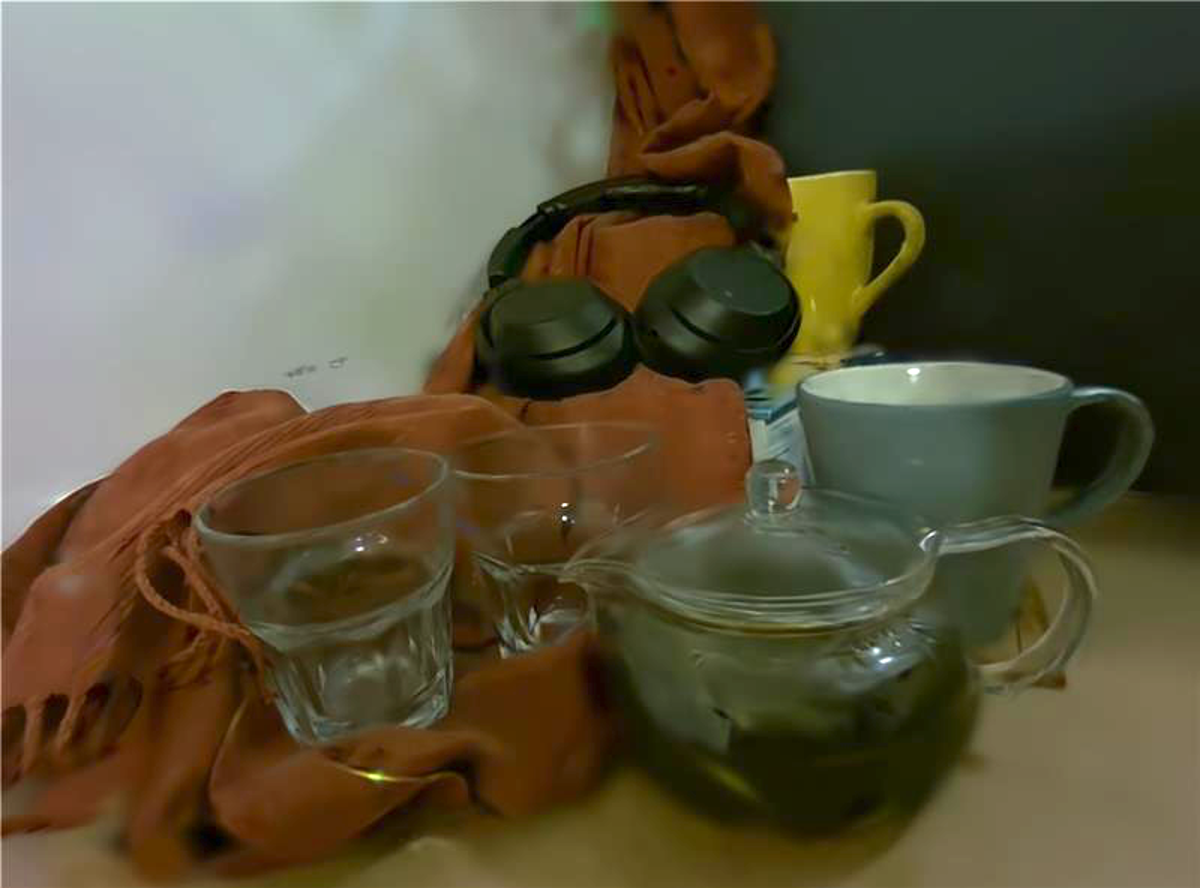}
  \caption{
\textbf{Self-captured Scenes:} Challenging scenes and objects, for example, transparent glasses and crinkled texture of salad are captured by our monocular SLAM running live.}\label{fig:qualitative_real}
\end{figure}

We report both the 3D reconstruction of the SLAM dataset and self-captured sequences. In Fig.~\ref{fig:qualitative_tum}, we visualise the monocular SLAM reconstruction of fr1/desk.
The placements of the Gaussians are geometrically sensible and are 3D coherent, and our rendering from the different viewpoints highlights the quality of our systems' novel view synthesis.
In Fig.~\ref{fig:qualitative_real}, we self-capture challenging scenes for monocular SLAM. By not explicitly modelling a surface, our system naturally handles transparent objects which is challenging for many other SLAM systems.

%% file: sec/5_conclusion.tex
\section{Conclusion}

We have proposed the first SLAM method using 3D Gaussians as a SLAM representation. Via efficient volume rendering, our system significantly advances the fidelity and diversity of object materials a live SLAM system can capture.
Our system achieves state-of-the-art performance across benchmarks for both monocular and RGB-D cases.
Interesting directions for future research are the integration of loop closure for handling large-scale scenes and extraction of geometry such as surface normal as Gaussians do not explicitly represent the surface.

%% file: sec/6_acknowledgement.tex
\section{Acknowledgement}
Research presented in this paper has been supported by
Dyson Technology Ltd. We are very grateful to Eric Dexheimer, Kirill Mazur, Xin Kong, Marwan Taher, Ignacio Alzugaray, Gwangbin Bae, Aalok Patwardhan, and members of
the Dyson Robotics Lab for their advice and insightful discussions.

%% file: sec/X_suppl.tex
\clearpage
\setcounter{page}{1}

{\begin{flushleft}\LARGE \textbf{Supplementary Material} \end{flushleft}}

\section{Implementation Details}
\subsection{System Details and Hyperparameters}
\subsubsection{Tracking and Mapping (Sec. \ref{sec:tracking} and \ref{sec:mapping})}
\paragraph{Learning Rates}
We use the Adam optimiser for both camera poses and Gaussian parameters optimisation.
For camera poses, we used 0.003 for rotation and 0.001 for translation. For 3D Gaussians, we used the default learning parameters of the original Gaussian Splatting implementation~\cite{kerbl3Dgaussians}, apart from in monocular setting where we increase the learning rate of the positions of the Gaussians $\meanW$ by a factor of 10.

\paragraph{Iteration numbers} 100 tracking iterations are performed per frame for across all experiments. However, we terminate the iterations early if the magnitude of the pose update becomes less than $10^{-4}$.
For mapping, 150 iterations are used for the single-process implementation. 

\paragraph{Loss Weights}
Given a depth observation, for tracking we minimise both photometric Eq.~\eqref{eqn:photometric} and geometric residual Eq.~\eqref{eqn:geometric_residual} as:
\begin{equation}
    \min_{\camCW \in \SE{3}} \lambda_{pho} E_{pho} + (1-\lambda_{pho}) E_{geo}~,
\end{equation}
and similarly, for mapping we modify Eq.~\eqref{eqn:mapping} to:
\begin{align}
    \min_{\substack{\camCW^k \in \SE{3}, \gaussians, \\ \forall k \in \window}}
    \sum_{\forall k \in \window}&(\lambda_{pho} E^{k}_{pho}+ (1-\lambda_{pho}) E^k_{geo}) \nonumber \\&+ \lambda_{iso} E_{iso}
    ~.
\end{align}
We set $\lambda_{pho} = 0.9$ for all RGB-D experiments, and
$\lambda_{iso} = 10$ for both monocular and RGB-D experiments. 



\subsubsection{Keyframing (Sec.~\ref{sec:keyframing})}
\paragraph{Gaussian Covisibility Check (Sec.~\ref{sec:gaussian_covisibility})}
As described in Sec.~\ref{sec:keyframing}, keyframe selection is based on the covisibility of the Gaussians. Between two keyframes $i$, $j$, we define the covisibility using the Intersection of Union (IOU) and Overlap Coefficient (OC):
\begin{align}
IOU_{cov}(i, j) &= \frac{| \gaussians^v_i \cap  \gaussians^v_j|}{| \gaussians^v_i \cup  \gaussians^v_j |}~, \\
OC_{cov}(i, j) &= \frac{| \gaussians^v_i \cap  \gaussians^v_j|}{\min(|\gaussians^v_i|,  |\gaussians^v_j|)}~,
\end{align}
where $\gaussians^v_i$ is the Gaussians visible in keyframe $i$, based on visibility check described in Section~\ref{sec:gaussian_covisibility}, Gaussian Covisibility.
A keyframe $i$ is added to the keyframe window $\kfwindow$ if given last keyframe $j$, $IOU_{cov}(i, j) < kf_{cov}$ or if the relative translation $t_{ij} > kf_m \hat{{D}}_i$, where  $\hat{{D}}_i$ is the median depth of frame $i$.
For Replica $kf_{cov} = 0.95, kf_m = 0.04$ and for TUM $kf_{cov} = 0.90, kf_m = 0.08$.
We remove the registered keyframe $j$ in $\kfwindow$ if the $OC_{cov}(i, j) < kf_{c}$, where keyframe $i$ is the latest added keyframe. For both Replica and TUM, we set the cutoff to $kf_{c} = 0.3$.
We set the size of the keyframe window to be for Replica, $|\kfwindow| = 10$, and for TUM, $|\kfwindow| = 8$.

\paragraph{Gaussian Insertion and Pruning (Sec. \ref{sec:GaussianInsertionPruning})}
As we optimise the positions of Gaussians and prune geometrically unstable Gaussians, we do not require any strong prior such as depth observation for Gaussian initialisation.
When \textbf{inserting} new Gaussians in a monocular setting, we randomly sample the Gaussians position $\meanW$ using rendered depth $D$. 
Since the estimated depth may sometimes be incorrect, we account for this by initialising the Gaussians with some variance.
For a pixel $p$ where the rendered depth $\mathcal{D}_p$ exists, we sample the depth from $\mathcal{N} (\mathcal{D}_p, 0.2 \sigma_D)$. Otherwise, for unobserved regions, we initialise the Gaussians by sampling from $\mathcal{N} (\hat{D}, 0.5 \sigma_D)$, where $\hat{D}$ is the median of $D$. 
For \textbf{pruning}, as described in Section~\ref{sec:GaussianInsertionPruning}, we perform visibility-based pruning, where if new Gaussians inserted within the last 3 keyframes are not observed by at least 3 other frames, they are pruned. We only perform visibility-based pruning once the keyframe window $\kfwindow$ is full. Additionally, we prune all Gaussians with opacity of less than 0.7.

\section{Evaluation details}
\subsection{Camera Tracking Accuracy (Table~\ref{table:ate_tum_rgbd} and Table~\ref{tab:ate_replica_rgbd})}
\subsubsection{Evaluation Metric}
We measured the keyframe absolute trajectory error (ATE) RMSE. For monocular evaluation, we perform scale alignment between the estimated scale-free and ground-truth trajectories. For RGB-D evaluation, we only align the estimated trajectory and ground truth without scale adjustment.

\subsubsection{Baseline Results}
\paragraph{Table~\ref{table:ate_tum_rgbd}} Numbers for monocular DROID-SLAM~\cite{Teed:Deng:NIPS2021} and ORB-SLAM~\cite{Mur-Artal:etal:TRO2017} is taken from \cite{li2023dense}. We have locally run DSO~\cite{Engel:etal:PAMI2017}, DepthCov~\cite{Dexheimer:etal:CVPR2023} and DROID-VO~\cite{Teed:Deng:NIPS2021} -- which is DROID-SLAM without loop closure and global bundle adjustment.
For the RGB-D case, numbers for NICE-SLAM~\cite{Zhu2022CVPR}, DI-Fusion~\cite{huang2021difusion}, Vox-Fusion~\cite{yang2022vox}, Point-SLAM~\cite{Sandström2023ICCV} are taken from Point-SLAM~\cite{Sandström2023ICCV}, and numbers for iMAP~\cite{Sucar:etal:ICCV2021}, BAD-SLAM~\cite{badslam}, Kintinous~\cite{Whelan:etal:IJRR2015}, ORB-SLAM~\cite{Mur-Artal:etal:TRO2017} are from iMAP~\cite{Sucar:etal:ICCV2021}, and ald all the other baselines: ESLAM~\cite{johari-et-al-2023}, Co-SLAM~\cite{wang2023coslam} are from each individual papers.
\paragraph{Table~\ref{tab:ate_replica_rgbd} and~\ref{table:rendering}}
We took the numbers from Point-SLAM~\cite{Sandström2023ICCV} paper.
\paragraph{Table~\ref{table:time_memory}}
The numbers are from Co-SLAM~\cite{wang2023coslam} paper.

\subsection{Rendering Performance (Table~\ref{table:rendering})}
\renewcommand{\arraystretch}{1.1}
\begin{table*}
\centering
 \resizebox{\linewidth}{!}{
\begin{tabular}{ccccccccccc||c}
\hline
\textbf{Method} & \textbf{Metric} & \textbf{room0} & \textbf{room1} & \textbf{room2} & \textbf{office0} & \textbf{office1} & \textbf{office2} & \textbf{office3} & \textbf{office4} & \textbf{Avg.} & \textbf{Rendering FPS} \\
\hline
\multirow{3}{*}{NICE-SLAM~\cite{Zhu2022CVPR}} & PSNR[dB] ↑ & 22.12 & 22.47 & 24.52 & 29.07 & 30.34 & 19.66 & 22.23 & 24.94 & 24.42& \multirow{3}{*}{0.54}  \\
 & SSIM ↑ & 0.689 & 0.757 & 0.814 & 0.874 & 0.886 & 0.797 & 0.801 & 0.856 & 0.809&   \\ 
 & LPIPS↓ & 0.33 & 0.271 & 0.208 & 0.229 & 0.181 & 0.235 & 0.209 & 0.198 & 0.233&   \\ \hline
 
\multirow{3}{*}{Vox-Fusion~\cite{yang2022vox}} & PSNR[dB] ↑ & 22.39 & 22.36 & 23.92 & 27.79 & 29.83 & 20.33 & 23.47 & 25.21 & 24.41&  \multirow{3}{*}{\underline{2.17}}  \\
 & SSIM ↑ & 0.683 & 0.751 & 0.798 & 0.857 & 0.876 & 0.794 & 0.803 & 0.847 & 0.801&  \\ 
 & LPIPS↓ & 0.303 & 0.269 & 0.234 & 0.241 & 0.184 & 0.243 & 0.213 & 0.199 & 0.236&   \\ \hline
 
\multirow{3}{*}{\begin{tabular}{c} Point-SLAM~\cite{Sandström2023ICCV} \end{tabular}} & PSNR[dB] ↑ & \underline{32.40} & \underline{34.08} & \underline{35.5} & \underline{38.26} & \underline{39.16} & \underline{33.99} & \underline{33.48} & \underline{33.49} & \underline{35.17}&  \multirow{3}{*}{1.33} \\ 
 & SSIM ↑ & \textbf{0.974} & \textbf{0.977} & \textbf{0.982} & \textbf{0.983} & \textbf{0.986} & \underline{0.96} & \underline{0.960} & \textbf{0.979} & \textbf{0.975}&   \\ 
 & LPIPS↓ & \underline{0.113} & \underline{0.116} & \underline{0.111} & \underline{0.1} & \underline{0.118} & \underline{0.156} & \underline{0.132} & \underline{0.142} & \underline{0.124}&   \\ \hline
\multirow{3}{*}{\bf{Ours}} & PSNR[dB] ↑ & \textbf{34.83} & \textbf{36.43} & \textbf{37.49} & \textbf{39.95} & \textbf{42.09} & \textbf{36.24} &\textbf{36.7} & \textbf{36.07} & \textbf{37.50}&  \multirow{3}{*}{\textbf{769}}\\
 & SSIM ↑ & \underline{0.954} & \underline{0.959} & \underline{0.965} &  \underline{0.971} & \underline{0.977} & \textbf{0.964} & \textbf{0.963} & \underline{0.957} & \underline{0.960}&  \\
 & LPIPS↓ & \textbf{0.068} & \textbf{0.076} & \textbf{0.075} & \textbf{0.072} & \textbf{0.055} & \textbf{0.078} & \textbf{0.065} & \textbf{0.099}  & \textbf{0.070}&   \\ \hline
\end{tabular}
}
\caption{\textbf{Rendering performance comparison of RGB-D SLAM methods on Replica.} Our method outperforms most of the rendering metrics compared to existing methods. Note that Point-SLAM uses sensor depth (ground-truth depth in Replica) to guide sampling along rays, which limits the rendering performance to existing views. 
The numbers for the baselines are taken from~\cite{Sandström2023ICCV}.}
\label{table:rendering_full}
\end{table*}

We provide the full detail of the rendering performance evaluation in Table~\ref{table:rendering_full}.

In Table~\ref{table:rendering}, we reported the photometric quality metrics (PSNR, SSIM and LPIPS) and rendering fps of our methods. We demonstrated that our rendering fps (769) is much higher than other existing methods (VoxFusion is the second best with 2.17fps). Here we describe the detail of how we measured the fps. 
The rendering time refers to the duration necessary for full-resolution rendering ($1200\times680$ for the Replica sequence). For each method, we perform 100 renderings and report the average time taken per rendering. The reported rendering fps is found by taking 1 and dividing it by the average rendering time.
We summarise the numbers in Table~\ref{table:rendering_fps}. Note that the ``rendering fps'' means the fps just for the forward rendering, which differs from the end-to-end system fps reported in Table~\ref{table:timing_TUM} and~\ref{table:timing_replica}.

\begin{table}[!t]
\centering
 \resizebox{\linewidth}{!}
 {
\begin{tabular}{ccc}
\hline
\ \textbf{Method} & \textbf{Rendering FPS ↑} & \textbf{\begin{tabular}{c}
Rendering time \\ 
per image [s] ↓
\end{tabular}} \\ \hline
   NICE-SLAM~\cite{Zhu2022CVPR}  & 0.54  &  1.85 \\ 
   Vox-Fusion~\cite{yang2022vox} & \underline{2.17}  &  \underline{0.46}  \\ 
   Point-SLAM~\cite{Sandström2023ICCV} & 1.33 & 0.75 \\
   Ours & \bf{769} & \bf{0.0013} \\
 \hline 
\end{tabular}
}
\caption{\textbf{Further detail of Rendering FPS and Rendering Time comparison based on Table~\ref{table:rendering}}.}
\label{table:rendering_fps}
\end{table}

\subsection{The convergence basin analysis (Table~\ref{table:conv_basin} and Fig~\ref{fig:conv_basin})}
\subsubsection{The detail of the benchmark Dataset}
\begin{figure}[!tbp]
  \center
  \includegraphics[width=\linewidth]{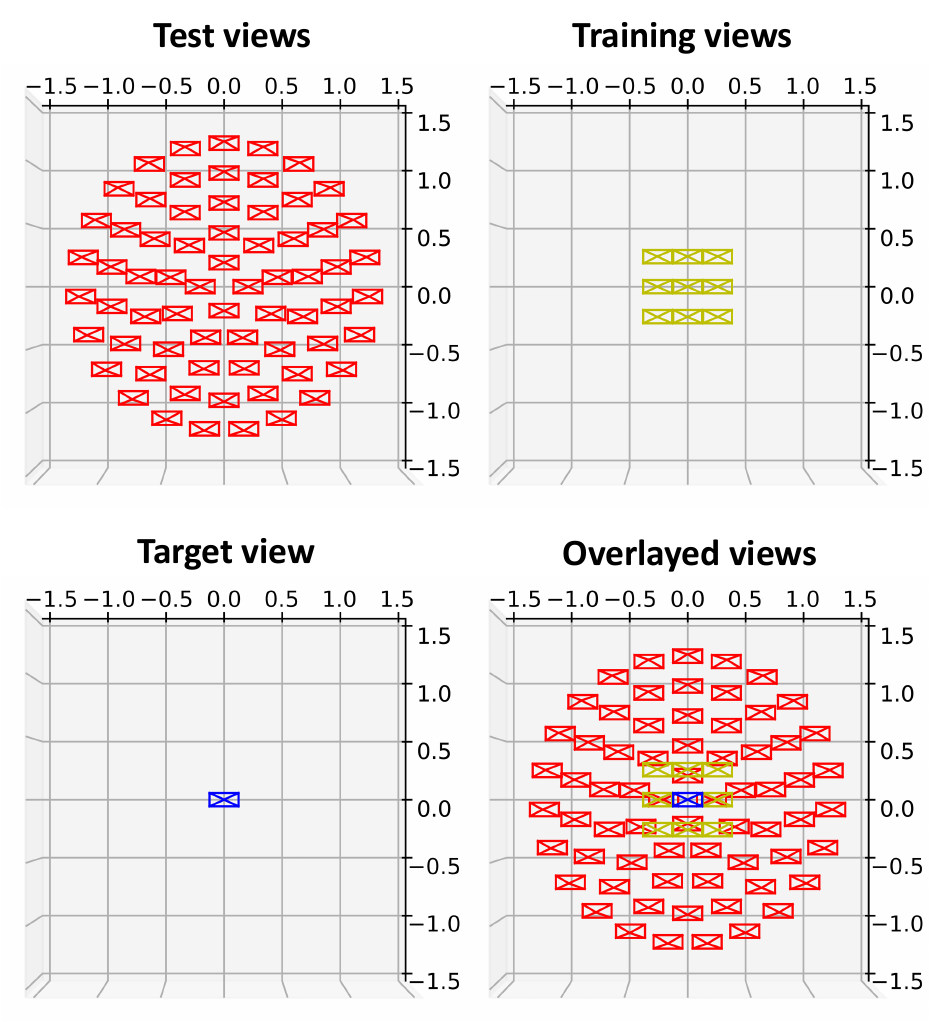}
  \caption{\textbf{2D Visualisation of the camera pose distributions used for convergence basin analysis in Figure~\ref{fig:conv_basin}.}   
   }
   \label{fig:conv_basin_vis}
\end{figure}

For convergence basin analysis, we create three datasets by rendering the synthetic Replica dataset.
In addition to the qualitative visualisation in Figure~\ref{fig:conv_basin}, we report more detailed camera pose distributions in Figure~\ref{fig:conv_basin_vis}.   
Figure~\ref{fig:conv_basin_vis} shows the camera view frustums of the test (red), training (yellow) and target (blue) views. As we mentioned in the main paper, we set the training view in the shape of a square with a width of 0.5m and test views are distributed with radii ranging from 0.2m to 1.2m, covering a larger area than the training views. We only apply displacements to the camera translation but not to the rotation. For each sequence, we use a total of 67 test views.

\subsubsection{Training setup}
For each method, the 3D representation is trained for 30000 iterations using the training views. Here, we detail the training setup of each of the methods:
\paragraph{Ours}
We evaluated our method under two settings: ``w/ depth'' and ``w/o depth'', where we train the initial 3D  Gaussian map  ${\gaussians}_{init}$ with and without depth supervision.
In the ``w/o depth'' setting, the 3D Gaussians' positions are randomly initialised, and we minimise the monocular mapping cost Eq.~\eqref{eqn:mapping} for the 3D Gaussian training, but keeping the camera poses fixed. Specifically, let $k\in\mathbb{N}$ be a number of training views and 3D Gaussians $\gaussians$, we find ${\gaussians}_{init}$ by: 
\begin{equation}
    \gaussians_{init} = 
    \argmin_{\substack{\gaussians}}
    \sum_{\forall k \in \window} E^{k}_{pho}+ \lambda_{iso} E_{iso}~.
\end{equation}\label{eq:ginit_mono}
Note that training views' camera poses $\camCW^{k}$ are fixed during the optimisation.

In the ``w/ depth'' setting, we train the Gaussian map by minimising the same cost function as our RGB-D SLAM system:

\begin{align}
    \gaussians_{init} = 
    \argmin_{\gaussians}
    \sum_{\forall k \in \window} (\lambda_{pho} E^{k}_{pho} &+ (1-\lambda_{pho}) E^{k}_{geo}) \nonumber \\ 
    &+  \lambda_{iso} E_{iso}~,  \label{eq:ginit_rgbd}
\end{align}
where we use $\lambda_{pho}=0.9$ and $\lambda_{iso}=10$ for all the experiments

\paragraph{Baseline Methods}
For Hash Grid SDF, we trained the same network architecture as Co-SLAM~\cite{wang2023coslam}. For MLP SDF, we trained the network of iMAP~\cite{Sucar:etal:ICCV2021}. For both baselines, we supervised networks with the same loss functions as Co-SLAM, which are colour rendering loss $L_{rgb}$, depth rendering loss $L_{depth}$, SDF loss $L_{fs}$, free-space loss $L_{fs}$, and smoothness loss $L_{smooth}$. Please refer to the original Co-SLAM paper for the exact formulation (equation (6) - (9)). All the training hyperparameters (e.g. learning rate of the network, number of sampling points, loss weight) are the same as Co-SLAM's default configuration of the Replica dataset. While Co-SLAM stores training view information by downsampling the colour and depth images, we store the full pixel information because the number of training views is small.

\subsubsection{Testing Setup}
For testing, we localise the camera pose by minimising only the photometric error against the ground-truth colour image of the target view.

\paragraph{Ours} Let the camera pose $\camCW \in \SE{3}$ and initial 3D Gaussians ${\gaussians}_{init}$, the localised camera pose $\camCW^{est}$ is found by:
\begin{equation}
     \camCW^{est} = \argmin_{\substack{\camCW}} \left\| I(\gaussians_{init}, \camCW) - {\gtimage}_{target} \right\|_1~.
\end{equation}
Note that ${\gaussians}_{init}$ is fixed during the optimisation. We initialise $\camCW$ at one of the test view's positions, and optimisation is performed for 1000 iterations. We perform this localisation process for all the test views and measure the success rate. Camera localisation is successful if the estimated pose converges to within 1cm of the target view within the 1000 iterations. 

\paragraph{Baseline Methods} For the baseline methods, the camera localisation is performed by minimising colour volume rendering loss $L_{rgb}$, while all the other trainable network parameters are fixed. The learning rates of the pose optimiser are also the same as Co-SLAM's default configuration of Replica dataset.


\begin{table}[!t]
\centering
 {
\begin{tabular}{cccc}
\hline
\ \textbf{Method} & \textbf{Total Time [s]} & \textbf{FPS}& \\ \hline
   Monocular & 798.9 & 3.2  \\ 
   RGB-D   &   986.7 & 2.5 \\ 
 \hline 
\end{tabular}
}
\caption{\textbf{Performance Analysis using fr3/office.} Both monocular and RGB-D implementations use multiprocessing. We report \textbf{the total execution time of our system}, FPS computed by dividing the total number of processed frames by the total time.}\label{table:timing_TUM}
\end{table}

\begin{table}[!t]
\centering
 {
\begin{tabular}{cccc}
\hline
\ \textbf{Method} & \textbf{Total Time [s]} & \textbf{FPS} & \\ \hline
   RGB-D  &  1111.1& 1.8   \\ 
   RGB-D (sp) & 1904.7  & 1.1  \\ 
 \hline 
\end{tabular}
}
\caption{\textbf{Performance Analysis using replica/office1.} RGB-D uses a multi-process implementation and RGB-D-sp is the single-process implementation. We report \textbf{the total execution time of our system}, FPS computed by dividing the total number of processed frames by the total time.}\label{table:timing_replica}
\vspace{2em}

\end{table}

\section{Further Ablation Analysis (Table~\ref{table:ablation})}
\subsection{Pruning Ablation (Monocular input)}
In Table~\ref{tab:pruning_ablation}, we report the ablation study of our proposed Gaussian pruning, which prunes randomly initialised 3D Gaussians effectively in a monocular SLAM setting. As the result shows, Gaussian pruning plays a significant role in enhancing camera tracking performance. This improvement is primarily because, without pruning, randomly initialised Gaussians persist in the 3D space, potentially leading to incorrect initial geometry for other views.

\begin{table}[!t]
\centering
 \resizebox{\linewidth}{!}{
\begin{tabular}{cccccc}
\hline
\textbf{Input} & \textbf{Method} & \textbf{fr1/desk} & \textbf{fr2/xyz} & \textbf{fr3/office} & \textbf{Avg.} \\ \hline

 \multirow{2}{*}{Mono}
 & w/o pruning & 78.2 & 4.5   & 57.0 & 46.6 \\ 
     & \bf{Ours}   & \bf{3.78} & \bf{4.60} & \bf{3.50} & \bf{3.96} \\ 
 \hline 
\end{tabular}
}
\caption{\textbf{Pruning Ablation Study on TUM RGB-D dataset (Monocular Input).} Numbers are camera tracking error (ATE RMSE) in cm.}
\end{table}\label{tab:pruning_ablation}

\subsection{Isotropic Loss Ablation (RGB-D input)}
Table~\ref{tab:ablation_iso_tum} and~\ref{tab:ablation_iso_replica} report the ablation study of the effect of isotropic loss $E_{iso}$ for RGB-D input. In TUM, as Table~\ref{tab:ablation_iso_tum} shows, isotropic regularisation does not improve the performance but only shows a marginal difference. However, for Replica, as summarised in Table~\ref{tab:ablation_iso_replica}, isotropic loss significantly improves camera tracking performance. Even with the depth measurement, since rasterisation does not consider the elongation along the viewing axis. Isotropic regularisation is required to prevent the Gaussians from over-stretching, especially for textureless regions, which are common in Replica.

\begin{table}[!t]
\centering
 \resizebox{\linewidth}{!}{
\begin{tabular}{cccccc}
\hline
\textbf{Input} & \textbf{Method} & \textbf{fr1/desk} & \textbf{fr2/xyz} & \textbf{fr3/office} & \textbf{Avg.} \\ \hline

 \multirow{2}{*}{RGB-D}
 & w/o $E_{iso}$ & 1.60 & \bf{1.42}   & \bf{1.32} & \bf{1.43} \\ 
 & \bf{Ours}  &  \bf{1.50} & 1.44 & 1.49  & 1.47   \\ 
 \hline 
\end{tabular}
}

\caption{\textbf{Isotropic Loss Ablation Study on TUM RGB-D dataset (RGB-D input).}  Numbers are camera tracking error (ATE RMSE) in cm.}
\label{tab:ablation_iso_tum}
\end{table}

\begin{table}[!t]
\centering
 \resizebox{\linewidth}{!}{
\begin{tabular}{ccccccccccc}
\hline
\textbf{Method} & \textbf{r0} & \textbf{r1} & \textbf{r2} & \textbf{o0} & \textbf{o1} & \textbf{o2} & \textbf{o3} & \textbf{o4} & \textbf{Avg.} \\ \hline
\begin{tabular}{c}
     w/o $E_{iso}$
\end{tabular} &  \bf{0.44}         & 0.86         &  \bf{0.28}         & 0.75            & 0.99             & 0.36            & 0.28           & 2.6             & 0.82           \\ 

\bf{Ours} &  \bf{0.44}         &\bf{0.32}         & {0.31}         &  \bf{0.44}             & \bf{0.52}             & \bf{0.23}             & \bf{0.17}             & \bf{2.25}             & \bf{0.58}         \\

\hline
\end{tabular}
}
\caption{\textbf{Isotropic Loss Ablation Study on Replica dataset (RGB-D input).} Numbers are camera tracking error (ATE RMSE) in cm.}
\label{tab:ablation_iso_replica}
\end{table}

\subsection{Effect of Spherical Harmonics (SH)}
While we disabled SHs in the main paper for simplicity, here we report the ablation study of the effect of SHs. The 3DGS paper~\cite{kerbl3Dgaussians} shows that addition of SH leads to small improvements in rendering metrics, and we have found similar improvement with SH enabled in our system (Tab.\ref{tab:tum_rendering_metrics}a). We did not observe a significant change in runtime with SH enabled, but it notably increases Gaussian map size and hence GPU memory usage.
Though an analytical Jacobian propagates the gradients from SH to camera poses, ATE marginally gets worse when SH is enabled (Tab.~\ref{tab:tum_memory}), as SH may incorrectly explain non-view directional effects caused by the camera motion, degrading the trajectory estimate.

\subsection{Mapping Performance with ORB-SLAM}
One of the most straightforward approaches for real-time operation is to combine an existing tracking system and 3DGS. In particular, frame-based SLAM methods have been well-studied for years and 
is capable of providing reliable tracking.
In this section, we compare our unified 3DGS-based method to the combined approach. We have run RGB-D ORB-SLAM to recover the poses and train 3DGS with the poses and sensor depth of the keyframes, equivalent to performing offline splatting. Though ORB-SLAM is best in terms of ATE (Tab.1 main), we find no significant difference across the rendering metrics (Tab.\ref{tab:tum_rendering_metrics}b). 
SH is omitted in the synthetic Replica dataset as it contains no view-directional effects.
While using an off-the-shelf tracker with a 3DGS mapper is possible,  this work has focused on the value of the 3DGS throughout the entire algorithms,
which is unexplored and therefore provides new insights. Further performance improvement of the unified approach will be an interesting future work.

\subsection{Large-scale Scenes with Stereo Inputs:}
This work focuses on pioneering 3DGS-based SLAM for live operation in small-scale scenes. However, we tested our method on the large-scale EuRoC Machine Hall dataset with depth from stereo (Tab.\ref{tab:euroc}). Fig.1 is a qualitative reconstruction result from our system.
Our method is competitive in ``easy'' sequences, although performance drops for more difficult, longer sequences.
Note that Point-SLAM~\cite{Sandström2023ICCV} fails on all sequences in this dataset.
In future work, we expect to improve our method by incorporating loop closure.
In principle, loop closure will be easier to incorporate compared to other representations such as voxel grids (where feature allocations are fixed), via a method similar to surfel-based approaches like ElasticFusion~\cite{Whelan:etal:RSS2015}.

\begin{table}[!t]
\center  
\includegraphics[width=\linewidth]{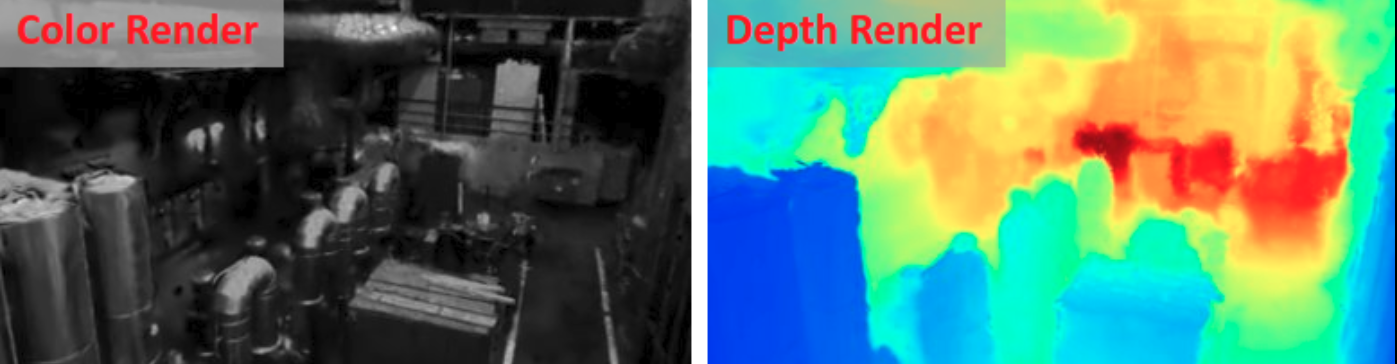}
\centering
 \resizebox{\linewidth}{!}{
\begin{tabular}{cccccc}
\hline
& \textbf{01-easy} & \textbf{02-easy} & \textbf{03-medium} & \textbf{04-difficult} & \textbf{05-difficult}  \\ \hline
Point-SLAM~\cite{Sandström2023ICCV} & - & - & - & - & - \\
Ours & \dashuline{0.121}         &  \dashuline{0.141}        & 2.197                  & 4.515             & 3.190                         \\ 
Vins-Fusion~\cite{qin2019b}      & 0.540         & 0.460         & \dashuline{0.330}         & \dashuline{0.780}             & \dashuline{0.500}                                      \\
SVO~\cite{Forster:etal:ICRA2014} & \underline{0.040}         &  \underline{0.070}        & \underline{0.270}                  & \underline{0.170}             & \underline{0.120}                         \\ 
ORB-SLAM3~\cite{Campos:etal:TRO2021} & \bf{0.029}         & \bf{0.019}         & \bf{0.024}         & \bf{0.085}            & \bf{0.052}                                       \\ 
\hline
\end{tabular}
}
\caption{ATE RMSE (meter) on EuRoC Machine Hall with Stereo Depth. Baseline numbers of classical methods are from~\cite{Campos:etal:TRO2021}. The third best result is highlighted with a dash line.}
\label{tab:euroc}
\end{table}

\begin{table}[!t]
\centering
\resizebox{\linewidth}{!}{
\begin{tabular}{cccccccccc}
\hline
&& \multicolumn{3}{c}{TUM} & \multicolumn{3}{c}{Replica} \\
& Method & PSNR $\uparrow$ & SSIM  $\uparrow$& LPIPS  $\downarrow$ & PSNR $\uparrow$ & SSIM  $\uparrow$& LPIPS  $\downarrow$\\
\hline
 \multirow{3}{*}{(a)} &Ours (w/o SH)   &  21.89 & 0.733  & 0.327  & 38.94 & 0.968 & 0.0703 \\
 &Ours (w. SH)    &  24.37 & 0.804 & 0.225  &  - & - & - \\
 &Point-SLAM      &  21.39 &	0.727 &	0.463  &  24.37 & 0.840 & 0.185  \\
\hline
\multirow{2}{*}{(b)} &ORB+GS (w/o SH)     &  25.12 & 0.837 & 0.161  &  37.11 & 0.964 & 0.040 \\
& ORB+GS (w.SH)    &  25.44 & 0.842 & 0.146  & - & - & -  \\
\hline
\end{tabular}
}
\caption{Mean Rendering metrics for TUM and Replica (RGBD). 
}
\label{tab:tum_rendering_metrics}
\end{table}

\begin{table}[!t]
\centering
\resizebox{\linewidth}{!}{
\begin{tabular}{ccccc||cc}
\hline
\multicolumn{5}{c||}{\bf{Memory Usage for RGB-D SLAM}} &\multicolumn{2}{c}{\bf{ATE RMSE}}  \\
\hline
\begin{tabular}{c} Ours \\ (w/o SH) \end{tabular} & 
\begin{tabular}{c} Ours \\ (w. SH) \end{tabular} &
Point-SLAM  & 
\begin{tabular}{c} ORB+GS \\ (w/o SH) \end{tabular} & 
\begin{tabular}{c} ORB+GS \\ (w. SH) \end{tabular} & 
\begin{tabular}{c} Ours \\ (w/o SH) \end{tabular} & 
\begin{tabular}{c} Ours \\ (w. SH) \end{tabular} \\
3.97MB & 11.47MB &  38.0MB & 45.97MB  & 186.5MB & 1.47cm& 1.56cm \\ 
\hline 
\end{tabular}
}
\caption{Mean Memory and ATE metrics for TUM (RGBD).}
\label{tab:tum_memory}
\end{table}

\subsection{Memory Consumption and Frame Rate (Table.~\ref{table:time_memory})}

\subsubsection{Memory Analysis}
In memory consumption analysis, for Table.~\ref{table:time_memory}, we measure the final size of the created Gaussians. The memory footprint of our system is lower than the original Gaussian Splatting, which uses approximately 300-700MB for the standard novel view synthesis benchmark dataset~\cite{kerbl3Dgaussians}, as we only maintain well-constrained Gaussians via pruning and do not store the spherical harmonics.

\subsubsection{Timing Analysis}
To analyse the processing time of our monocular/RGB-D SLAM system, we measure the total time required to process all frames in the TUM-RGBD fr3/office dataset. This approach assesses the performance of our system as a whole, rather than isolating individual components. By adopting this approach, we gain a more realistic understanding of the system's true performance which better reflects the real-world operating conditions, as it avoids the assumption of an idealised, sequential interleaving of the tracking and mapping processes. As shown in Table~\ref{table:timing_replica}, our system operates at 3.2 FPS with monocular and 2.5 FPS with depth. The FPS is found by dividing the number of processed frames by the total time.
We conducted a similar analysis with the Replica dataset office2. Here, we compare the RGB-D method with and without multiprocessing. As expected, single-process implementation takes longer as it performs more mapping iterations.

\section{Camera Pose Jacobian}
Use of 3D Gaussian as a primitive and performing camera pose optimisation is discussed in~\cite{keselman2022fuzzy}; however, the method assumes a smaller number of Gaussians and is based on ray-intersection not splatting; hence, is not applicable to 3DGS.
While many applications of 3DGS exist, for example, dynamic tracking and 4D scene representation~\cite{luiten2023dynamic, wu20234d}, they assume offline application and require accurate camera position.
In contrast, we perform camera pose optimisation by deriving the minimal analytical Jacobians on Lie group, and for completeness, we provide the derivation of the Jacobians presented in Eq.~\eqref{eqn:grad_meanC_camCW_W_camCW}.
\begin{align}
\mpd{\meanC}{\camCW} 
    &= \lim_{\tauC \to 0}\frac{\Exp(\tauC) \cdot \meanC - \meanC}{\tauC} \\
    &= \lim_{\tauC \to 0}\frac{(\identity + \tauC^\wedge) \cdot  \meanC - \meanC}{\tauC} \\
    &= \lim_{\tauC \to 0}\frac{\tauC^\wedge \cdot  \meanC}{\tauC} \\
    &= \lim_{\tauC \to 0}\frac{\theta^{\times} \meanC + \rho}{\tauC}  \\
    &= \lim_{\tauC \to 0}\frac{-\meanC^{\times} \theta + \rho}{\tauC} \\
    &= \begin{bmatrix} \identity & - \meanC^\times\end{bmatrix}
\end{align}
where $\cam\cdot \mathbf{x}$ is the group action of $\cam \in \SE{3}$ on $\mathbf{x} \in \RR^3$.

Simiarly, we compute the Jacobian with respect to $\matW$. Since the translational component is not involved, we only consider the rotational part $\rotCW$ of $\camCW$.
\begin{align}
\mpd{\matW}{\rotCW} &=  \lim_{\thetaC \to 0}\frac{\Exp(\thetaC) \circ \matW - \matW}{\thetaC} \\
&= \lim_{\thetaC \to 0}\frac{(\identity + \thetaC^{\wedge}) \circ \matW - \matW }{\thetaC} \\
&= \lim_{\thetaC \to 0}\frac{\thetaC^{\wedge} }{\thetaC} \circ \matW\\
&= \lim_{\thetaC \to 0}\frac{\thetaC^\times }{\thetaC}\circ \matW
\end{align}
Since skew symmetric matrix is:
\begin{equation}
    \theta^{\times} = \begin{bmatrix}
        0 &-\thetaC_z & \thetaC_y\\
        \thetaC_z & 0 & -\thetaC_x\\
        -\thetaC_y & \thetaC_x & 0\\
    \end{bmatrix}
\end{equation}
The partial derivative of one of the component (e.g. $\thetaC_x$) is:
\begin{equation}
    \pd{\theta^\times}{\thetaC_x} = \begin{bmatrix}
        0 & 0 & 0\\
        0 & 0 & -1\\
        0 & 1 & 0\\
    \end{bmatrix} = \mathbf{e}_1^{\times}
\end{equation}
where $\mathbf{e}_1 = [1, 0, 0]^{\top}, \mathbf{e}_2 = [0, 1, 0]^{\top}, \mathbf{e}_3 = [0, 0, 1]^{\top}$.
\begin{equation}
\pd{\matW}{\theta_x} = \mathbf{e}_1^{\times} \matW = \begin{bmatrix}
    \mathbf{0}_{1\times3} \\
    -\matW_{3,:} \\
     \hphantom{-}\matW_{2,:}\end{bmatrix}\label{eqn:grad_W_theta_x}
\end{equation}
\begin{equation}
\pd{\matW}{\theta_y} = \mathbf{e}_2^{\times} \matW = \begin{bmatrix}
    \hphantom{-}\matW_{3,:} \\
     \mathbf{0}_{1\times3}  \\
     -\matW_{1,:}\end{bmatrix}\label{eqn:grad_W_theta_y}
\end{equation}
\begin{equation}
\pd{\matW}{\theta_z} = \mathbf{e}_3^{\times} \matW = \begin{bmatrix}
    -\matW_{2,:} \\
     \hphantom{-}\matW_{1,:} \\
     \mathbf{0}_{1\times3} 
     \end{bmatrix}\label{eqn:grad_W_theta_z}
\end{equation}
where $\matW_{i, :}$ refers to the $i$th row of the matrix.
After column-wise vectorisation of Eq.~\eqref{eqn:grad_W_theta_x},~\eqref{eqn:grad_W_theta_y},~\eqref{eqn:grad_W_theta_z}, and stacking horizontally we get:
\begin{align}
    \mpd{\matW}{\rotCW} = \begin{bmatrix}
    -\matW_{:, 1}^\times \\
    -\matW_{:, 2}^\times \\
    -\matW_{:, 3}^\times\\
    \end{bmatrix}
    ~,
\end{align}
where $\matW_{:, i}$ refers to the $i$th column of the matrix.
Since the translational part is all zeros, with this we get Eq.~\eqref{eqn:grad_meanC_camCW_W_camCW}.

\section{Additional Qualitative Results}
\begin{figure*}[!tbp]
  \center
  \includegraphics[width=\linewidth]{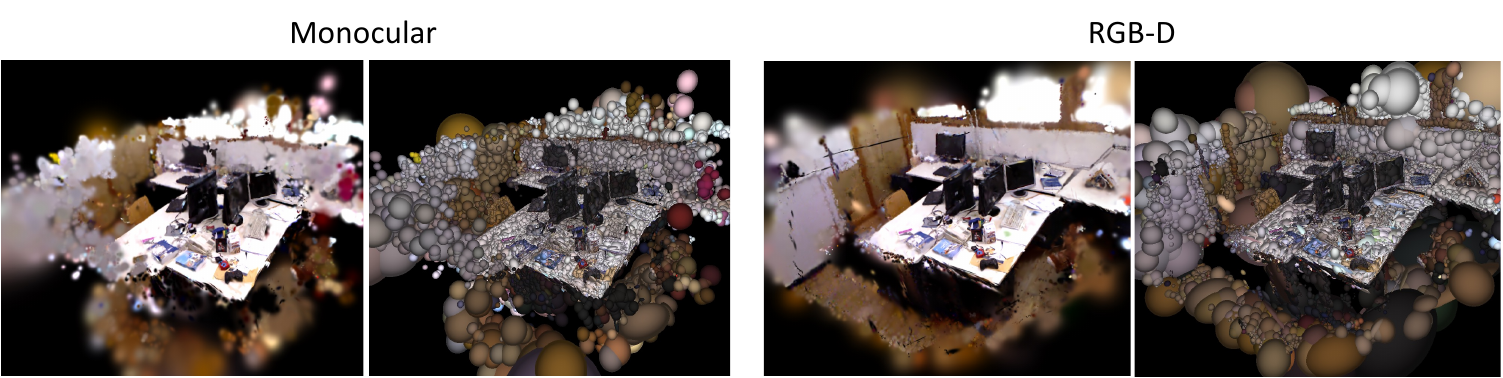}
  \caption{
  \textbf{Novel view rendering and Gaussian visualizations on TUM fr1/desk} 
  }\label{fig:fr1_map}
  
  \center
  \includegraphics[width=\linewidth]{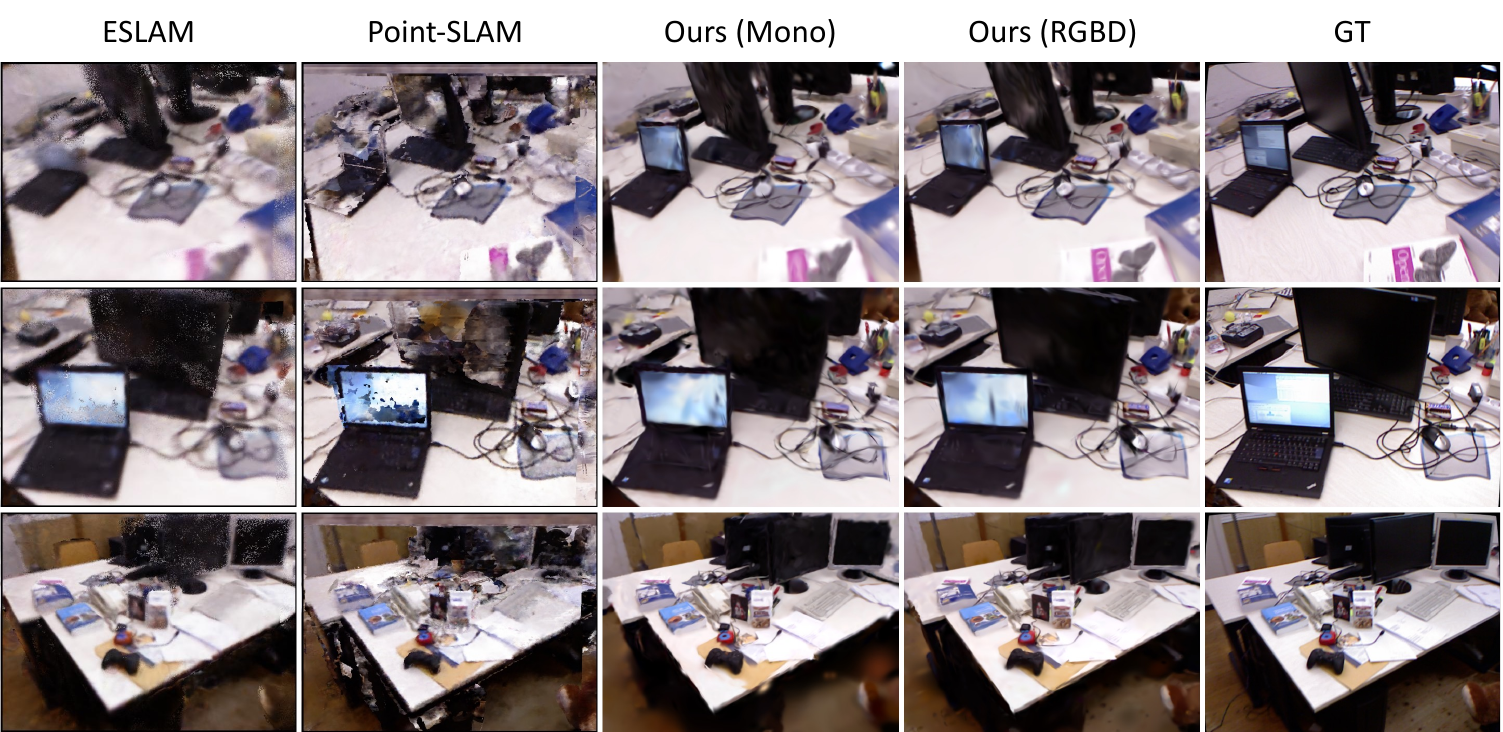}
  \caption{
  \textbf{Rendering comparison on TUM fr1/desk} 
  }\label{fig:fr1_rendering}
\end{figure*}

\begin{figure*}[!tbp]
  \center
  \includegraphics[width=\linewidth]{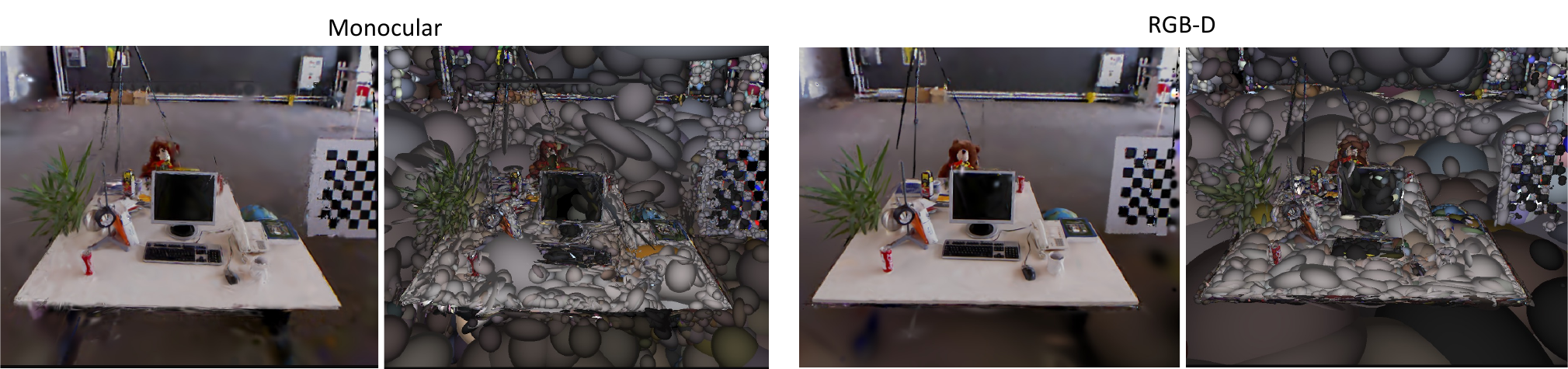}
  \caption{
  \textbf{Novel view rendering and Gaussian visualizations on TUM fr2/xyz} 
  }\label{fig:fr2_map}
  
  \center
  \includegraphics[width=\linewidth]{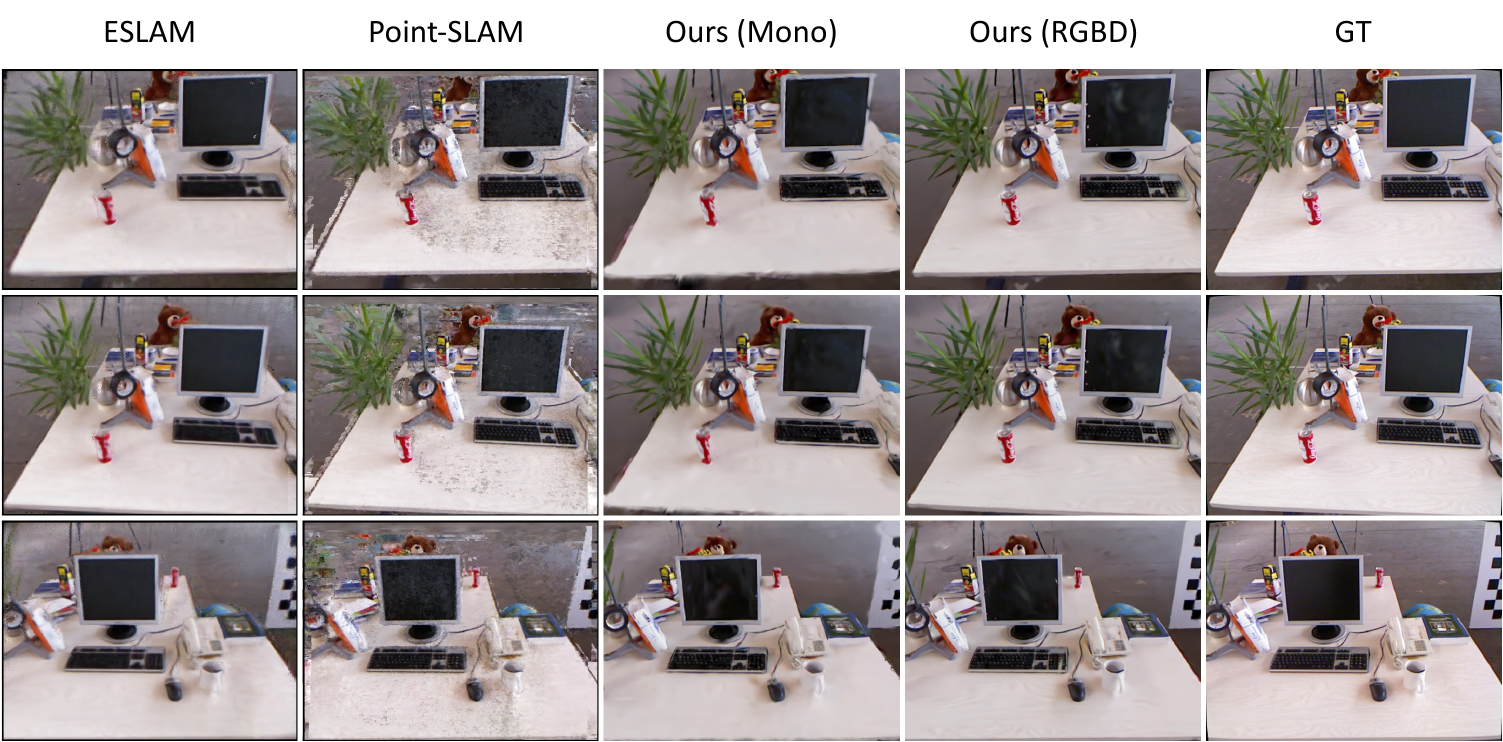}
  \caption{
  \textbf{Rendering comparison on TUM fr2/xyz} 
  }\label{fig:fr2_rendering}
\end{figure*}

\begin{figure*}[!tbp]
  \center
  \includegraphics[width=\linewidth]{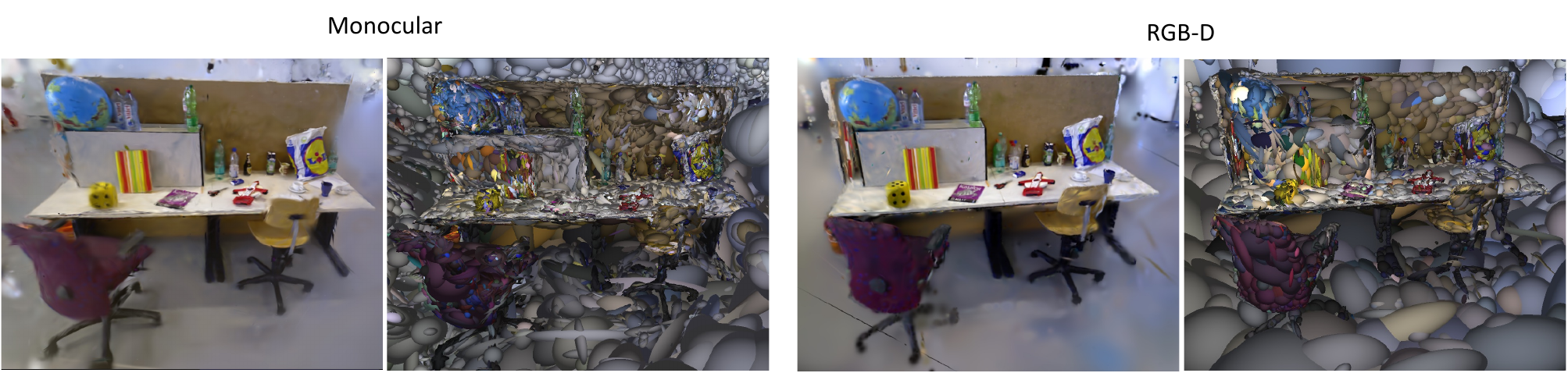}
  \caption{
  \textbf{Novel view rendering and Gaussian visualizations on TUM fr3/office} 
  }\label{fig:fr3_map}
  
  \center
  \includegraphics[width=\linewidth]{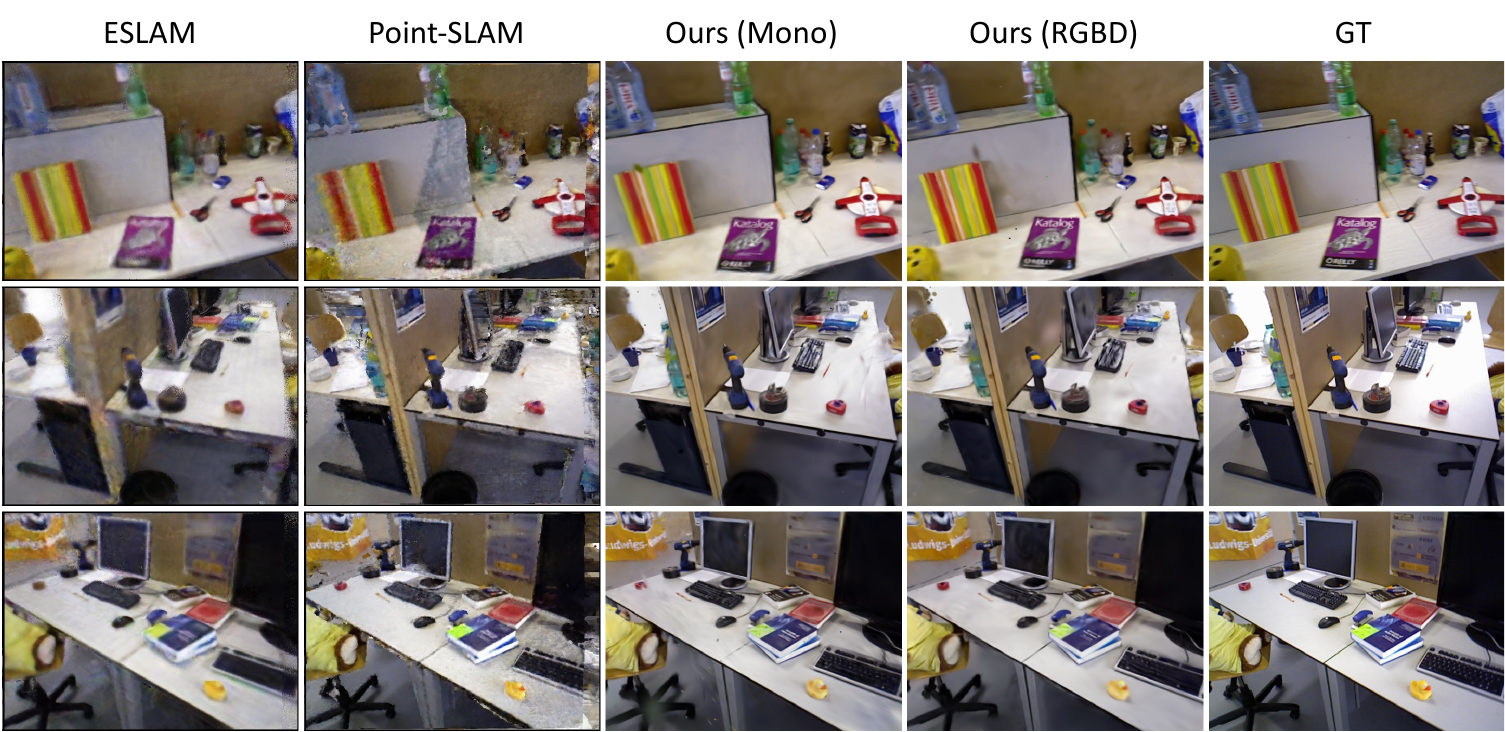}
  \caption{
  \textbf{Rendering comparison on TUM fr3/office} 
  }\label{fig:fr3_rendering}
\end{figure*}

\begin{figure*}[!tbp]
  \center
  \includegraphics[width=\linewidth]{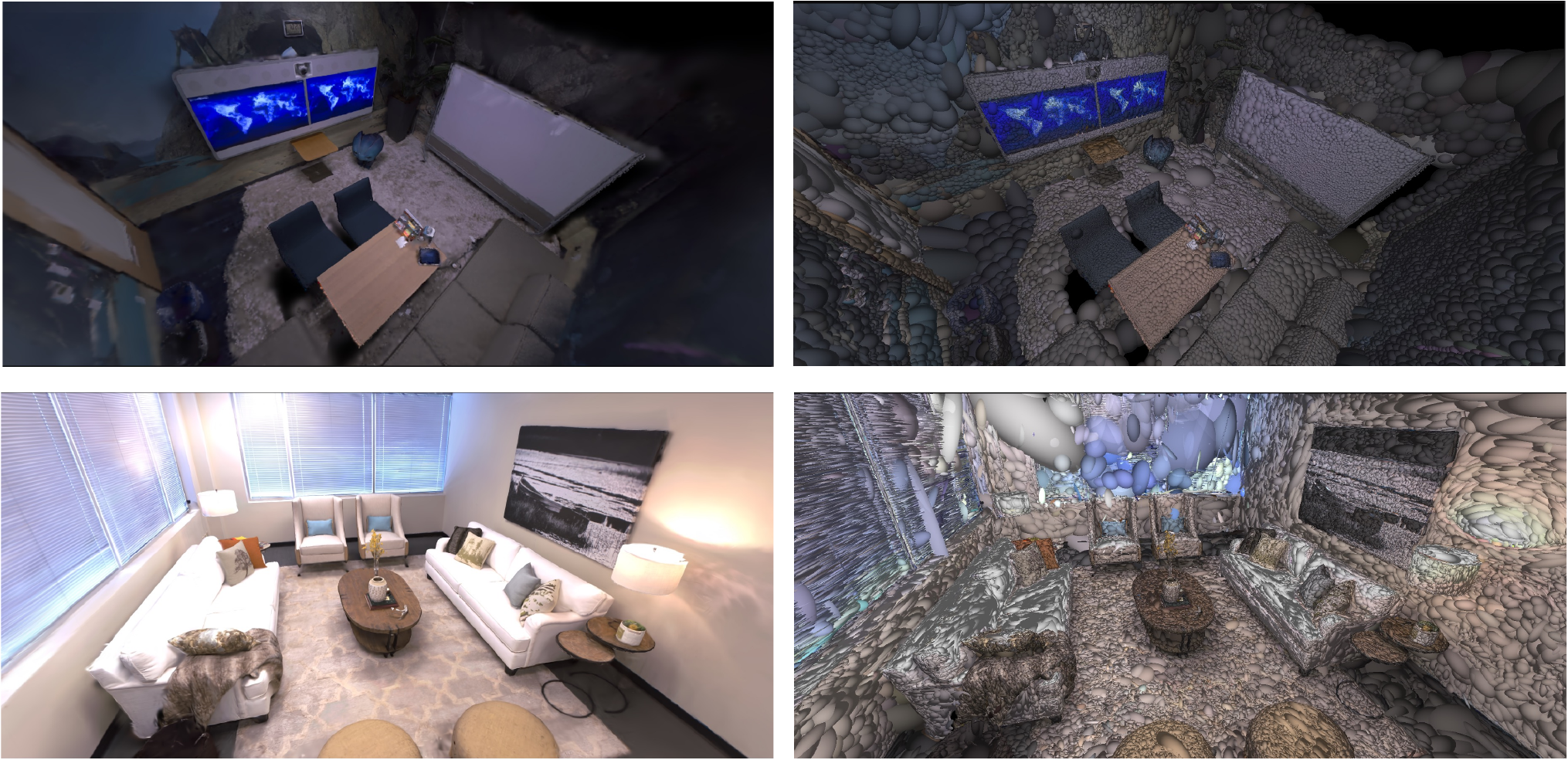}
  \caption{
  \textbf{Novel view rendering and Gaussian visualizations on Replica} 
  }\label{fig:replica_map}
  
  \center
  \includegraphics[width=\linewidth]{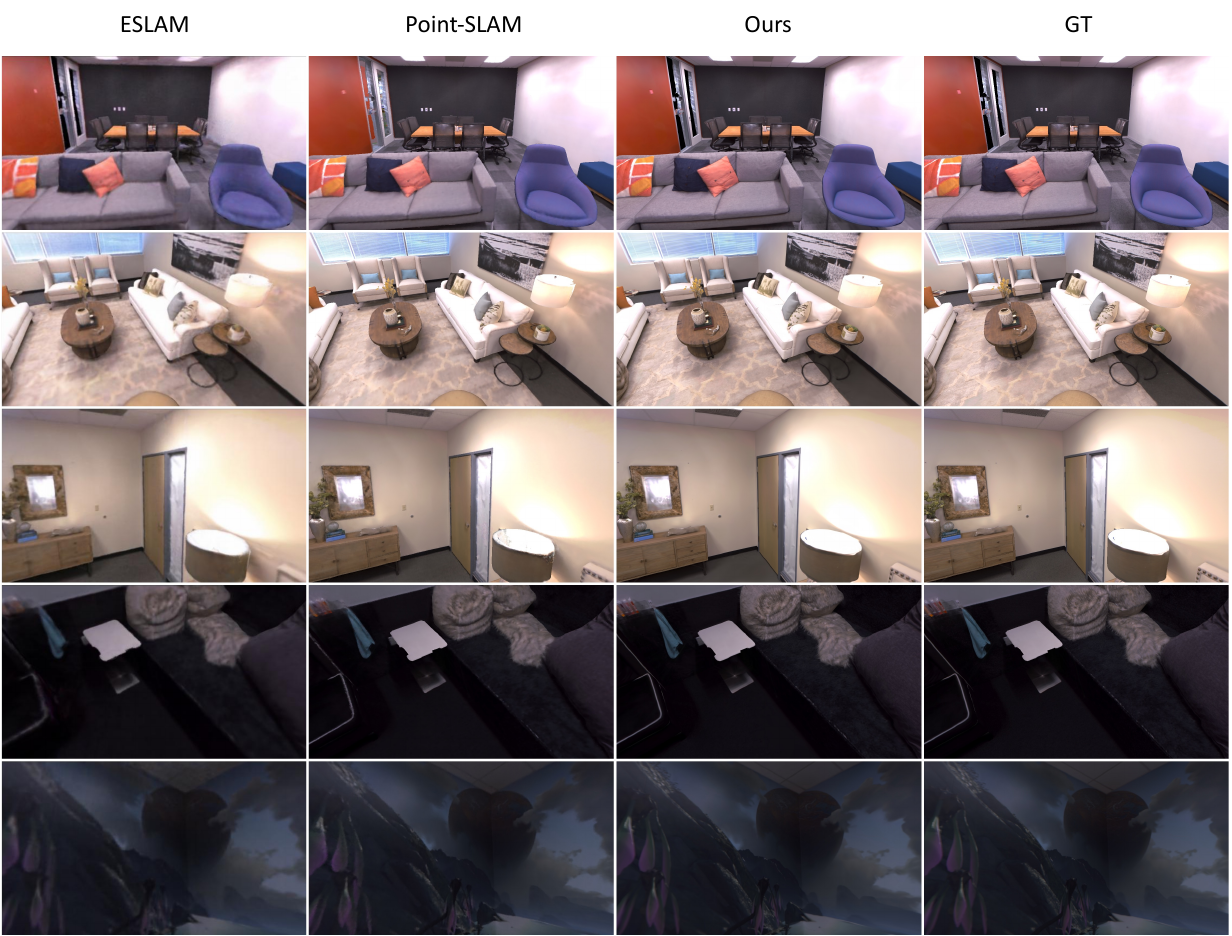}
  \caption{
  \textbf{Rendering comparison on Replica} 
      }\label{fig:replica_rendering}
\end{figure*}

We urge readers to view our supplementary video for convincing qualitative results.
In Fig.~\ref{fig:fr1_map} - Fig.~\ref{fig:replica_rendering}, we further show additional qualitative results. We visually compare other state-of-the-art SLAM methods using differentiable rendering (Point-SLAM~\cite{Sandström2023ICCV} and ESLAM~\cite{johari-et-al-2023}).

\section{Limitation of this work}
Although our novel Gaussian Splatting SLAM shows competitive performance on experimental results, the method also has several limitations. 
\begin{itemize}
    \item Currently, the proposed method is tested only on small room-scale scenes. For larger real-world scenes, the trajectory drift is inevitable. This could be addressed by integrating a loop closure module into our existing pipeline. 
    \item Although we achieve interactive live operation, hard real-time operation on the benchmark dataset (30 fps on TUM sequences) is not achieved in this work. To improve speed, exploring a second-order optimiser would be an interesting direction.
\end{itemize}

\newpage

\newpage

\newpage

